%% file: main.tex
\documentclass{article}

\usepackage{arxiv}

\usepackage[utf8]{inputenc} % allow utf-8 input
\usepackage[T1]{fontenc}    % use 8-bit T1 fonts
\usepackage{hyperref}       % hyperlinks
\usepackage{url}            % simple URL typesetting
\usepackage{booktabs}       % professional-quality tables
\usepackage{amsfonts}       % blackboard math symbols
\usepackage{nicefrac}       % compact symbols for 1/2, etc.
\usepackage{microtype}      % microtypography
\usepackage{lipsum}

\usepackage{graphicx}
\graphicspath{{figures/}}  % Location of 
\usepackage[table,xcdraw]{xcolor}
\usepackage{longtable}
\usepackage[normalem]{ulem}
\useunder{\uline}{\ul}{}
\usepackage{lscape}
\usepackage{longtable}
\usepackage{placeins}
\usepackage{amsmath}
\usepackage{subcaption}
\usepackage{amssymb}
\usepackage{float}
\title{A Spectral Enabled GAN for\\ Time Series Data Generation}

\author{
  Kaleb ~Smith\\
  Department of Computer Engineering and Sciences\\
  Florida Institute of Technology\\
  Melbourne, FL 32901 \\
  \texttt{ksmith012007@my.fit.edu} \\
   \And
  Anthony O.~Smith \\
  Department of Computer Engineering and Sciences\\
  Florida Institute of Technology\\
  Melbourne, FL 32901 \\
  \texttt{anthonysmith@fit.edu} \\
}

\begin{document}
\maketitle

\begin{abstract}
Time dependent data is a main source of information in today's data driven world. Generating this type of data though has shown its challenges and made it an interesting research area in the field of generative machine learning. One such approach was that by Smith et al. who developed Time Series Generative Adversarial Network (TSGAN) which showed promising performance in generating time dependent data and the ability of few shot generation though being flawed in certain aspects of training and learning. This paper looks to improve on the results from TSGAN and address those flaws by unifying the training of the independent networks in TSGAN and creating a dependency both in training and learning. This improvement, called unified TSGAN (uTSGAN) was tested and comapred both quantitatively and qualitatively to its predecessor on 70 benchmark time series data sets used in the community. uTSGAN showed to outperform TSGAN in 80\% of the data sets by the same number of training epochs and 60\% of the data sets in 3/4th the amount of training time or less while maintaining the few shot generation ability with better FID scores across those data sets. 
\end{abstract}

% keywords can be removed

\input{Introduction}
\input{RelatedWork}
\input{Approach}
\input{Experiments}
\input{Conclusion}

\bibliographystyle{unsrt}  


\bibliography{references.bib}

\end{document}

%% file: Introduction.tex
\section{Introduction}
\label{intro}

% Before advanced algorithms which parallel artificial intelligence
% (AI), the human race relied on, well, humans to complete many tasks
% that have now been successfully automated. One such task is data generation,
% where there is a desire to replicate \emph{real} data by creating
% \emph{fake} data. Hand-crafted data generation tasks were labor intensive
% and costly. For example, graphic artists rendering imaginary faces
% could take days or weeks to create one \emph{realistic} fake face.
% Another example is when entire systems and scenarios are invented
% to mimic a real event, just so sensors can be hooked up and data collections
% can be performed. What triggered the AI craze for helping these laborious
% tasks to become easier for humans was a sub-field of machine learning
% called deep learning. Deep learning
% consists of algorithms based off of the human brain called Neural
% Networks (NN) that are stacked in many different
% configuration manners. The success
% that deep learning shows compared to older approaches is due to essentially
% taking a lot of the feature learning out of the hand of the human
% and into the hand of the algorithm, having self taught networks on
% slews of data. What has hindered many advancements in deep learning
% applications, however, is the lack of data to train these networks,
% and hence the importance of data generation. 

{U}{ntil} the recent thrust in deep learning algorithms, and their continued success with many machine learning tasks, deep learning was thought to be for data generation. The work by Goodfellow et al. \cite{Goodfellow2014GenerativeNets} introduced the concept of a Generative Adversarial Networks (GAN) to the community. Since then, Goodfellow's paper has over 18000 citations with thousands of variations of GANs. It is worth noting that from all the GANs, there is only a fragment (less than 1\%) of the total GAN methods dealing with one dimensional (1D) continuous valued time series data; the vast majority of them deal with images (for the sake of comparison, two dimensional (2D) data). This observation shows two things: first, GANs are really good at generating image data; second, there is a gap in time series generation methods using GANs. We believe this is due to the more developed image-based community for deep learning, more standardized data sets, larger data sets, and competitions for these algorithms. Throughout this paper our reference to time series is with respect to 1D continuous time series data. 

Another sub-field of deep learning is few shot learning: the ability for an algorithm to learn its task with little training data. In the time series community there is again a lack of deep learning type data sets with millions of labeled samples publicly available, as noted by Dau et al. \cite{Dau2018TheArchive} in their paper addressing the community for a time series data archive. This cascades down to the few time series generation GANs that do exist, many use their own \emph{private} data sets which are large enough to train their
networks. This results in a lack of few shot approaches since publicly available large data sets are sparse. 

%% file: RelatedWork.tex
\section{Related Works}
\label{relwks}

Generating time series data is a task that has been analyzed for years to help in many forecasting and prediction algorithms \cite{Zhang2003TimeModel,Kim2005SyntheticGeneration,Baldi2003TheProblem,Ngoko2014SyntheticModels,Griffin1984SignalTransform,Pepe2005OnTechnique,Bogardi1988PracticalZone,Gers2002LearningNetworks,Hontoria2002GenerationPerceptron}. Since uTSGAN addresses this gap in time series generation with GANs, we focus this related works on deep approaches to generate the same type of data, as well as GANs that deal with different conditional inputs similar to uTSGAN. 

The past works using GANs on time series deal primarily with adopting the GAN framework and using recurrent neural networks (RNN) for the architecture of the generator and discriminator. Mogren \cite{Mogren2016C-RNN-GAN:Training} was one of the first authors to try time series generation with a continuous RNN-GAN (C-RNN-GAN). C-RNN-GAN uses long-short term memory (LSTM) networks for the generator and discriminator, taking advantage of their ability to learn sequential patterns over time.  C-RNN-GAN uses a semi-conditional input \cite{Odena2017ConditionalGans} for their generation, having the random vector input be also conditioned with the generated data learned from the previous time step of the RNN. 

Esteban et al. \cite{Esteban2017Real-valuedGans} worked on a similar method as Mogren, creating a Recurrent GAN (RGAN) and a Recurrent Conditional GAN (RCGAN).  These approaches by Esteban et al. looked to mature the C-RNN-GAN by eliminating the conditional dependency of the generated data from previous time steps, as well as show the ability to generate time series with just a random input vector. Ramponi et al. \cite{Ramponi2018T-cgan:Sampling} also worked on a conditional GAN called Time-Conditional GAN (T-CGAN) which utilized
Convolutional Neural Network (CNN) in its generator and discriminator.  This work wasn't for generating time series from random input noise, rather trying to augment data with irregular time sampling, but we find the work showed great success with this task enough to be brought into our literature review. Many other papers since have utilized these two GANs in a magnitude of applications based research, generating time series data in diverse domains as text \cite{Zhang2018Metagan:Learning}, finance \cite{Simonetto2018GeneratingTransactions}, biosignals \cite{Haradal2018BiosignalNetworks,Zhu2017UnpairedNetworks}, sensor \cite{Alzantot2017Sensegen:Generation}, smart grid data \cite{Zhang2018GenerativeGrids}, traffic \cite{Parthasarathy2020ControlledGANs}, and renewable scenarios \cite{Chen2018Model-FreeNetworks}. These works all have in common an
underlying theme that conditional input is the best route in these GANs for generating realistic time series data. 

Drawing inspiration from these works, uTSGAN uses conditioning for our time series generation, although we learn our conditioning instead of force feeding the networks. Different than these works is the choice of loss function for the networks. We use the WGAN loss  \cite{GANarjovsky2017wasserstein,Gulrajani2017ImprovedGANs} in our work for its ability to combat mode collapse, as well as proven ability in generating better data than other objective functions.  RNNs also have unstable training and vanishing gradients when being
trained for easier tasks such as classification. In relation to the difficulty of a regular RNN, Esteban et al. \cite{Esteban2017Real-valuedGans} cites that the WGAN objective is difficult for RNNs due to the 1-Lipshitz constraint being kept throughout training. For this reason and the success of Ramponi et al. \cite{Ramponi2018T-cgan:Sampling}, we stray away from RNN architectures and instead opt for a architecture similar to the Fully Convolutional Network (FCN) \cite{Wang2017TimeBaseline} which has shown high achieving accuracy rates in time series classification. We look further into the GAN time series community to see if other works inspire uTSGAN for further development. 

Though RNNs are the typical method for time series networks, there are other works that see the value in using the WGAN objective similar to ours. Smith et al. \cite{Smith2019TimeGAN} used a simple, off the shelf WGAN, but with a residual network as the generator and a FCN as the discriminator. This work showed an importance of being able to uniquely turn a 2D WGAN into a 1D WGAN and produce decent time series data. Hartmann et al. \cite{Hartmann2018EEG-GAN:Signals} developed a GAN dealing strictly with electroencephaographic (EEG) time series data and a WGAN loss dubbed EEG-GAN. This paper used similar architectures as Ramponi et al. \cite{Ramponi2018T-cgan:Sampling} with CNNs being used for their generator and discriminator, but did not have any conditioned input. They modified the WGAN loss by allowing the gradient penalty term to vary during training and using a training technique called progressive growing which allows the network to scale resolution during training \cite{Karras2017ProgressiveVariation}.  EEG-GAN shows considerable results for EEG data, however, many hyperparameter tweaking and ad hoc training techniques were done unique to their EEG dataset in order to generate data.

Donahue et al. \cite{Donahue2018AdversarialSynthesis} looked at generating raw audio clips with a GAN called WaveGAN in an attempt to produce sound effect generation. Their approach used a 1D version of a 2D Deep Convolutional GAN (DCGAN) \cite{Radford2015UnsupervisedNetworks} and trained using the WGAN loss citing the same advantages of better training and generation of data. DCGANs tend to leave checkerboard-like artifacts in their generated samples \cite{Odena2017ConditionalGans}, and when transferred to audio these artifacts come in the form of pitch noises that overlap other frequencies in the samples. To combat this, WaveGAN introduces a phase shuffle operation before the input to the discriminator. This in turns acts like adding noise to the input of the discriminator (both real and synthetic data), a technique to stabilize training by making the task more difficult for the discriminator and as a result allowing the generator to learn better \cite{Karras2019ANetworks,Karras2017ProgressiveVariation,Snderby2017ARESOLUTION}. 

Brophy et al. \cite{Brophy2019QuickGANs} looked at the time series generation differently with their WGAN inspired method for generating medical data . Instead of generating time series with a WGAN, the authors convert their time series data into an image and then utilize the WGAN's image generation ability to generate synthetic time series images. Those generated images are then mapped back to time series data in which they use for comparison to the original data. Though the method showed decent results, the ability to map the time series to and from an image is not well explained and further investigation is needed. 

Brophy et al. were not the only researchers who saw promise in using GANs to generate images and then map those back to time series. This is seen plenty in a different attribute of time series data, speech and music generation. Donahue et al. \cite{Donahue2018AdversarialSynthesis} in the same paper as WaveGAN introduced SpecGAN, a semi-invertible spectrogram generating GAN.  Citing the fact many breakthroughs in audio classification come from looking at spectrograms \cite{Sainath2015LearningCLDNNs,JongpilLee2017Sample-levelWaveforms}, they felt there could be some use generating audio from the same time-frequency space. SpecGAN uses a similar DCGAN and WGAN loss as WaveGAN, but instead of generating audio it generates spectrograms. These synthetic spectrograms then have their preprocessing steps inverted and the Griffin-Lim algorithm \cite{Griffin1984SignalTransform} is used to estimate phase from the spectrogram in order to produce their audio signal. This Griffin-Lim algorithm is used to predict discarded phase information when a signal is converted into a spectrogram with a STFT. It iteratively attempts to find the waveform whose STFT magnitude is closest to the generated spectrogram. Griffin-Lim requires complex information generated from the spectrogram when the preprocessing steps are inverted in order to find the signal match.

Donahue et al. also introduced GANSynth \cite{Engel2019GansS}, a generative model for synthesizing audio from modeling log magnitudes and instantaneous frequencies (IF) of the signal. In a similar fashion to SpecGAN, GANSynth generates log-magnitude spectrograms with either the phase information appended or IF. Unlike its closely related counterpart, GANSynth adds a conditional input of musical pitch to the generation model and the adds an auxiliary classifier to the discriminator in order to identify the pitch used. After the generation of either mixture occurs, the generated data is converted back to audio by performing the inverse STFT. To note, in all of the GANs dealing with this audio generation, the task is trained on a vast amount of data (millions of samples) and took several days for many of the networks to train. Our work agrees with Donahue et al. in which we see a benefit of generating time series data from the time-frequency domain due to its ability to carry vast amount of information about the signal not seen by the waveform itself. 

A more recent paper by Kumar et al. introduced a text-to-speech GAN called MelGAN \cite{Kumar2019MelGAN:Synthesis}. This model was part of a end to end approach where text sequences were fed into Seq2Seq (a text to mel-spectrogram network) producing a mel-spectrogram which was then used as the input into MelGAN to generate the synthetic audio clip. Though the task of text-to-speech is one out of scope of this paper, we do draw inspiration from MelGAN with their generator's ability to produce audio from the mel-frequency cepstrum coefficients (MFCC) generated by their network. MelGAN also showed comparison with other text-to-speech competitors including Google's WaveNet\cite{Oord2016WaveNet:Audiob}, WaveGlow \cite{WAVEGLOW:Corporation}, Tactron 2 \cite{Shen2018NaturalPredictions}, and ClairNet \cite{Ping2017ClariNetIn}.   showing comparable ability to these more complicated autoregressive deep learning models for this task. 

Yoon et al. introduced TimeGAN \cite{Yoon2019Time-seriesNetworks} time in a recent paper for continuous time series generation. The authors mix the unsupervised learning framework of a GAN and the supervised training of autoregressive models for control of TimeGAN's learning. This paper's introduction to a supervised loss allows for better temporal behavior of the generated time series data and their embedding network inside, alongside the GAN architecture allows for a lower-dimensional adversarial learning space. TimeGAN showed good results on two open source data sets and used unique metrics to categorize their success against competitors. 

The latest paper for time series generation comes from Smith et al. with their already mentioned time series GAN (TSGAN) \cite{Smith2019TimeGAN}. TSGAN takes the best from many of these papers above and tries to simplify the creation of time series generation to just unsupervised learning and tries to let the novel architecture carry the workload. Similar to the reasons we mention above for uTSGAN, TSGAN uses conditioning input for the generation of time series signals and does so in the time-frequency domain. Since uTSGAN is inspired from TSGAN we will go into more detail about the methodology of the architecture in the next section of this paper.

%% file: Approach.tex
%% LyX 2.3.4.2 created this file.  For more info, see http://www.lyx.org/.
%% Do not edit unless you really know what you are doing.

\section{Our Approach to uTSGAN}
\label{ourapp}

\begin{figure*}
\begin{centering}
\includegraphics[scale=0.55]{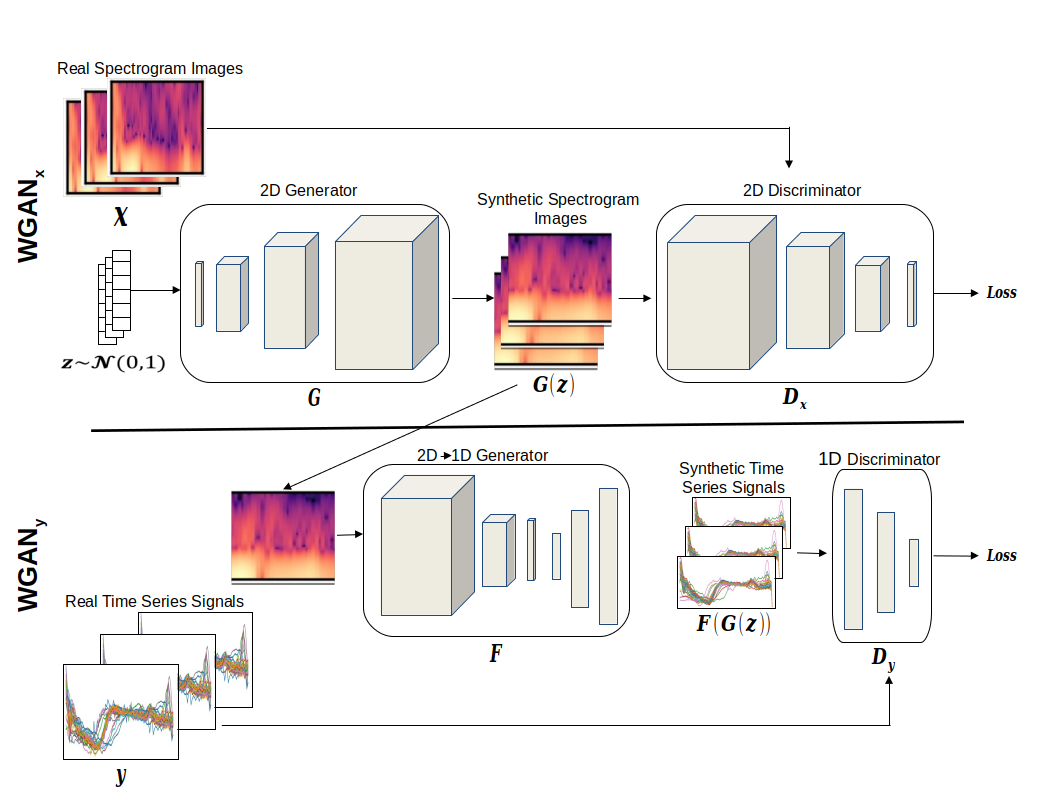}
\par\end{centering}
\caption{TSGAN architecture, two separate WGANs with two seperate tasks in their generation.}
\label{TSGAN_Architecture}
\end{figure*}

A brief review of TSGAN needs to be done to understand where uTSGAN is making contributions. TSGAN is a novel GAN architecture that uses two separate Wasserstein GANs (WGAN) in order to achieve its goal in time series generation. Note, for simplicity, here on out when we refer to WGAN we refer to the WGAN with gradient penalty (WGAN-GP) by Gulrajani et al.\cite{Gulrajani2017ImprovedGANs} and all WGAN notation is that of WGAN-GP. The TSGAN architecture can be seen in Fig \ref{TSGAN_Architecture} showing the two seperate WGANs and their inputs/outputs. TSGAN starts by generating synthetic spectrogram images in $WGAN_{x}$ and feeds those images to $WGAN_{y}$. Again, it is important to note the difference here compared to the other spectrogram related papers in Section \ref{relwks}.

In the papers by Donahue et al. \cite{Donahue2018AdversarialSynthesis} and Kumar et al. \cite{Kumar2019MelGAN:Synthesis} a visual representation of the signal was used to train their GAN. They also cited the ability for GANs to generate images extremely well compared to its efforts in time series generation, and the authors wanted to leverage this in their networks. In both papers, their the
underlying spectral information was used for training and then used thereafter to achieve the time series generation. For example, Donahue et al. in SpecGAN produces spectrogram images similar to TSGAN, but they must convert their spectrogram image back to its linear-amplitude magnitude spectra (done by inverse their preprocessing steps) and then estimate the time series by another algorithm offline (Griffin-Lim) in order to produce their time series data. 

This is not the case for TSGAN; instead of using spectra information directly, it relies on the image capture of the spectra in order to produce its time series. What does this mean? In essence, TSGAN takes an image of RGB values, representing the spectral information of the signal, and uses that information to infer and generate what time series signal created that spectrogram. This method eliminates the need for inverse preprocessing steps and does not rely on any underlying information which is needed when taking the STFT; i.e. window size, frequency, window type, hop length, etc. TSGAN is essentially a step in the right direction for a universal time series generation method, not dependent on the data being trained on or sub-characteristics of the data, i.e. frequency, collection methods, etc.

Where TSGAN faults though is two fold: first, its training is slow due to training two individual WGANs in serial to one another; second, there is no dependency on the two networks to perform optimally to assist one another in generation. Both of these issues are addressed in uTSGAN below and show how a simple combining the loss functions of $WGAN_{x}$ and $WGAN_{y}$ can generate better time series signals. 

Our goal is to unify the TSGAN architecture by creating a single loss function that can be optimized over the entire system. This conversion to a single loss function will bring a faster/better convergence for training and help produce better generated samples in the 1D domain. We look to build our new loss function and method by adapting the work of Smith et al. \cite{Smith2019TimeGAN} TSGAN's dual independent loss functions. First we will review TSGAN's original loss functions, then we will look at other works in the community which unify multiple generator architectures and multiple discriminator architectures to help us draw inspiration and credibility to create a single function describing two separate GANs.  Lastly we will show the new loss function holds the same properties as each of the independent loss functions in the original paper by Smith et al. \cite{Smith2019TimeGAN}. 

\[
WGAN_x =\mathcal{\mathbb{E}}_{x\sim\mathbb{P}_{x}}[D_{x}(x)]-\mathbb{E}_{z\sim\mathcal{N}(0,1)}[D_{x}(G(z))]
\]
\[
GP_x =\lambda_{x}\mathbb{E}_{\hat{x}\sim\mathbb{P}_{\hat{x}}}[(||\nabla_{\hat{x}}D_{x}(\hat{x})||_{2}-1)^{2}]
\]
\begin{equation}
L_{x}(G,D_{x},x,z)=\min_{G}\max_{D_{x}} WGAN_x + GP_x
\label{eq:L_x loss function}
\end{equation}

% \begin{equation}
% L_{x}(G,D_{x},x,z)=\min_{G}\max_{D_{x}}\mathcal{\mathbb{E}}_{x\sim\mathbb{P}_{x}}[D_{x}(x)]-\mathbb{E}_{z\sim\mathcal{N}(0,1)}[D_{x}(G(z))]+\lambda_{x}\mathbb{E}_{\hat{x}\sim\mathbb{P}_{\hat{x}}}[(||\nabla_{\hat{x}}D_{x}(\hat{x})||_{2}-1)^{2}]\label{eq:L_x loss function}
% \end{equation}

\[
WGAN_y =\mathcal{\mathbb{E}}_{y\sim\mathbb{P}_{y}}[D_{y}(y)]-\mathbb{E}_{z\sim\mathcal{N}(0,1)}[D_{y}(F(G(z)))]
\]
\[
GP_y =\lambda_{y}\mathbb{E}_{\hat{y}\sim\mathbb{P}_{\hat{y}}}[(||\nabla_{\hat{y}}D_{y}(\hat{y})||_{2}-1)^{2}]
\]
\begin{equation}
L_{y}(F,G,D_{y},y,z)=\min_{F}\max_{D_{y}} WGAN_y + GP_y
\label{eq:L_y loss function}
\end{equation}

% \begin{equation}
% L_{y}(F,G,D_{y},y,z)=\min_{F}\max_{D_{y}}\mathcal{\mathbb{E}}_{y\sim\mathbb{P}_{y}}[D_{y}(y)]-\mathbb{E}_{z\sim\mathcal{N}(0,1)}[D_{y}(F(G(z)))]+\lambda_{y}\mathbb{E}_{\hat{y}\sim\mathbb{P}_{\hat{y}}}[(||\nabla_{\hat{y}}D_{y}(\hat{y})||_{2}-1)^{2}]\label{eq: L_y loss function}
% \end{equation}

TSGAN's original architecture was modeled by two independent WGAN with gradient penalties, $WGAN_{x}$ and $WGAN_{y}$, both with their
own loss function, Eq \ref{eq:L_x loss function} and Eq \ref{eq:L_y loss function}, trained independently. Here we denote the first WGAN's 2D generator and discriminator as $G$ and $D_{x}$, the second WGAN 's 1D generator and discriminator as $F$ and $D_{y}$, the random sampled vector as $z$, the 2D spectrogram image data as $x,$ and the 1D time series data at $y$. Also important to note are the gradient penalty terms in Eq \ref{eq:L_x loss function} and Eq \ref{eq: L_y loss function}, which are the same gradient penalties from the original WGAN-GP paper by Gukrajani et al. \cite{Gulrajani2017ImprovedGANs} constraining the gradients to be norm 1 and keep each WGAN 1-Lipshitz continuous. We will maintain these notions throughout this section and the rest of the paper. The two independent loss function TSGANs showed to perform well in generating time series data, however there are several drawbacks to this approach.

First, computationally this way is more expensive than just a single loss function to train. The method requires the training of $WGAN_{x}$ and $WGAN_{y}$ independently in serial which adds double the computational requirements to generate a single sample, $F(G(z)),$ of time dependent data. This serial training is due to the architecture setup where the conditioned input to the 1D generator, $F$, is only used after $WGAN_{x}$ has been trained for that iteration. Second, with the original TSGAN setup, $WGAN_{y}$ is completely dependent and conditioned on $WGAN_{x}$'s output; however in two independent loss functions there is no way to constrain the output of $WGAN_{x}$ to generate good samples to feed into $WGAN_{y}$. By combining the two loss functions, this dependency should be better conditioned during optimization by having having $F(G(z))$ be purely impacted by $G(z)$; essentially penalizing itself whenever a bad 2D generated sample,$G(z)$, constitutes a poor 1D generated sample $F(G(z))$. Ideally we look to merge the two loss functions together in order to show this relationship of the quality from $WGAN_{x}$'s 2D output to $WGAN_{y}$'s 1D output. 

To achieve this desired loss function we draw some conclusions from other works in the community that look to train GANs with multiple generators or multiple discriminators in their architecture. Hoang et al \cite{Enerators2018MGANWITH} created a system called Mixture Generative Adversarial Networks (MGAN) in which they show a system with multiple generators and one discriminator. The authors proposed using multiple generator networks inside MGAN in order to combat mode collapse in the generated samples. MGAN's multiple generators learn multiple distributions and are then averaged in order to cover a larger, more diverse distribution covering different data modes. 
\[
L_{MAD}=\min_{G_{1:K},C}\max_{D}\mathbb{E}_{\boldsymbol{x}\sim P_{data}}log[D(x)]\]
\[+\mathbb{E}_{\boldsymbol{x}\sim P_{model}}[log(1-D(\boldsymbol{x}))]\]
\begin{equation}
-\beta\{\sum_{k=1}^{K}\pi_{k}\mathbb{E}_{\boldsymbol{x}\sim P_{G_{K}}}[logC_{k}(\boldsymbol{x})]\}\label{eq:MAD_loss}
\end{equation}

\noindent They set up their system using one discriminator, $D$, some $K$ generators, and one classifier, $C$, which classifies which generator
the generated sample came from. Looking at the loss function in Eq \ref{eq:MAD_loss} we see the typical interaction between generators
and discriminators with the first two terms, being that of the loss of a vanilla GAN from Goodfellow et al. \cite{Goodfellow2014GenerativeNets}.  The last term is a softmax portion of a multiclassification problem, i.e. the classification from which generator did the sample. In there we see they sum the expectations of each of the $K$ generators and use this as their technique to pull together their generator outputs, but simple addition. This gives us cause to look into linearly adding our generator outputs as well since uTSGAN contains two separate generators. Where uTSGAN differs though is it contains two discriminators as well, while MGAN only has one.

Albuquerque et al. \cite{Jo2019Multi-objectiveDiscriminators} looked into the case of GANs with multiple discriminators in their paper dealing solely with how to work optimization techniques around them. The authors show there are four types of optimization strategies when dealing with multiple loss functions by looking at a method of multi-objective optimization by \cite{DebUsingAlgorithms} called Multiple Gradient Decent (MGD) \cite{Peitz2018Gradient-basedUncertainties,Paris2012Multiple-gradientMultiobjectif}. In MGD, you treat some $K$ number of loss functions, $l(\boldsymbol{x})$, all independently working on some sample vector in the variable space $\boldsymbol{x}$ and define the new function we wish to optimize as $\boldsymbol{L}(\boldsymbol{x})=[l_{1}(\boldsymbol{x}),l_{2}(\boldsymbol{x})...l_{K}(\boldsymbol{x})]^{T}$.  MGD will only work if $\boldsymbol{L}(\boldsymbol{x})$ is assumed to be a convex, continuously differentiable and smooth $l_{k}(\boldsymbol{x})\;\forall k\in[1,2,...,K]$.  MGD finds a common descent direction for $\boldsymbol{L}(\boldsymbol{x})$ by defining the convex hull the gradients of the loss functions, $\nabla l_{k}(\boldsymbol{x})$ and finding the minimum norm element within it. 

That norm element being what is used to update the gradient decent direction for the next iteration. Though MGD shows promise as the
better optimization strategy for GANs in terms of quality of generated samples, Albuquerque et al. point out that this method is very computationally expensive when multiple discriminators are used in a GAN structure. For our case in an improved TSGAN loss function, we take this computational time into consideration, but we realize with only two independent loss functions we could still compute rather quickly. The reason we decide to choose another method rather than MGD is for the sake of unifying the loss functions instead of simply solving the optimization problem. We would rather have a single loss function showing the dependencies between $WGAN_{x}$ and $WGAN_{y}$ and not have their loss function be treated independently any longer. 

Next, Albuqurque et al. look at methods tailored specifically around multiple discriminator GANs: Generative Multi-Adversarial Networks (GMAN) with its weighted average \cite{Durugkar2017GN} and a method called average loss minimization \cite{Neyshabur2017StabilizingProjections} . To note, Albuqurque et al. propose their own method for optimization of multiple discriminators called hypervolume maximization \cite{Jo2019Multi-objectiveDiscriminators} but similar to MGD it requires the loss functions to be treated independently and to use MGD for optimization. This is not the desired outcome for our new TSGAN loss function and we decide not to explore this method further. 

Average loss minimization and GMAN's weighted average are similar techniques to merge multiple loss functions. The approach by Durugkar
et al. in GMAN consists of training some $K$ number of unique discriminators and taking a softmax weighted arithmetic average of their outputs to train a single generator. The author's loss function in Eq \ref{eq:GMAN_Loss} now represents the entire system's loss as a linearly combined combination of all the discriminator's loss,

\begin{equation}
L=\sum_{k=1}^{K}\alpha_{k}L_{D_{k}},\label{eq:GMAN_Loss}
\end{equation}

\noindent where $\alpha_{k}$ is the softmax weighting function for each of the $K$ discriminators and $L_{D_{k}}$ is the standard GAN loss function per discriminator. The idea of this softmax weighting is developed because the discriminators had different architectures that cause there to be some really strong discriminators and really weak ones (i.e. being able to tell the difference between generated and real samples easily for strong discriminators and vice versa for weak).  An issue with many GANs is vanishing gradients due to weak gradients from strong discriminators and not allowing the generator to learn effectively and cause mode collapse. This method allows the generator to see the stronger gradients from the weaker discriminators in order to update and learn better. Durugkar et al. ran extensive experiments where the weighting scheme was changed from focusing on those weak discriminators and those strong and found that the simple average of discriminator\textquoteright s losses provided the best values for both metrics in most of the considered cases quality of generated samples. 

This result echoes to the proposed method by Neyshabur et al. of just a simple average loss minimization across the $K$ different discriminators all of the same architecture. The idea behind their method is discriminator, $D_{k}$, sees a different randomly projected lower dimensional version, $\boldsymbol{z}$, of the input image in order to better train the generator, $G$, on these conditional inputs. The loss function, Eq \ref{eq: averageLoss}, in this work by Neyshabur et al looks similar to Eq \ref{eq:GMAN_Loss} except now instead of a softmax the mean is taken throughout the discriminators. 

\begin{equation}
L=\sum_{k=1}^{K}\mathbb{E}_{\boldsymbol{z}\sim P_{z}}logD_{k}(G(\boldsymbol{z}))\label{eq: averageLoss}
\end{equation}
\noindent High level, the discriminative tasks in the projected space are harder which means the discriminators have a harder time discriminating between real and fake samples. This makes the discriminators avoid early convergence and learning which in turn battles mode collapse in the generated samples \cite{Goodfellow2014GenerativeNets}. We decide collectively from these two main results from Durugkar and Neyshabur, as well as the result from Hung et al, that we will use an average loss between our two network's discriminators. Since our main goal is to merge the independent loss functions in TSGAN together to unify the learning and dependency between $WGAN_{x}$ and $WGAN_{y}$, using this simple average indeed combines the two losses and then unifies the learning into one loss function. This means that our new unified loss function between $WGAN_{x}$ from Eq \ref{eq:L_x loss function} and $WGAN_{y}$ from Eq \ref{eq:L_y loss function} now becomes:

\begin{equation}
L_{TSGAN}(G,F,D_{x,}D_{y},x,y,z)=\frac{1}{2}[L_{x}+L_{y}].\label{eq:TGAN_Improved}
\end{equation}

\noindent The techniques described in the papers by Neyshabur et al. and Durugkar et al. deal primarily with the vanilla GAN loss function which has many problems with training and behaves poorly \cite{Goodfellow2014GenerativeNets,Salimans2016ImprovedGans}. Since we are dealing with WGANs in our to deal with this training problems and improve stability of our uTSGAN, we must make sure that the new loss function can still be considered a WGAN loss function. In other words, can we confirm we maintain similar attributes to the original WGAN papers by Arjousky et al. and Gulranjani et al. that has brought WGANs into such popularity in the community? We know WGANs are Lipshitz constrained and continuous which makes it possible for our gradient descent to work on optimization.  So we ask the question, are the sum of two Lipshitz continuous functions still Lipshitz continuous? We know continuity is preserved through the summation of continuous functions under the triangle inequality . The
Lipshitz constraint we know from Gulranjani et al. is Lipshitz continuous from the constraint on the gradient by penalizing the gradients to have norm at most one everywhere. So referring back to Eq \ref{eq:L_x loss function} and Eq \ref{eq:L_y loss function} we will continue to denote them as $L_{x}$ and $L_{y}$ that are Lipshitz continuous functions. We also know $L_{x}:\mathbb{R\rightarrow\mathbb{R}}$ and $L_{y}:\mathbb{R}\rightarrow\mathbb{R}$, $x,y\in\mathbb{R}$, and denote constants $C_{x}$and $C_{y}$ as the Lipshitz constants for $L_{x}$ and $L_{y}$ respectively. Then:
\[
|(L_{x}+L_{y})(x)-(L_{x}+L_{y})(y)|=
\]
\[|L_{x}(x)+L_{y}(x)-L_{x}(y)-L_{y}(y)|
\]
\[
=|L_{x}(x)-L_{x}(y)+L_{y}(x)-L_{y}(y)|\]
\[\leq|L_{x}(x)-L_{x}(y)|+|L_{y}(x)-L_{y}(y)|
\]
\[
\leq C_{x}|x-y|+C_{y}|x-y|=(C_{x}+C_{y})|x-y|,
\]

\noindent so $L_{x}+L_{y}$is Lipshitz constrained with constant $C=C_{x}+C_{y}$. By this, we can confirm $L_{TSGAN}$ is Lipshitz constrained and continuous and hence maintains the properties of WGANs. This is what we desire with our new loss function for our uTSGAN and training should be stabilized similar to that of the WGAN.  We still train the discriminator with a training ratio to the generator's single training update in order to preserve the works from the TSGAN and WGAN papers. With this new single loss function we cut the training time down simply by only having two discriminators update together instead of two discriminators updating in serial. We now use our uTSGAN to conduct similar experiments to the original TSGAN paper, showing qualitatively and quantitatively how uTSGAN improves time series generation. 

%% file: Experiments.tex
\section{Experiments and Results}
\label{expres}

\subsection{Data\label{subsec:Data}}

Like the original TSGAN paper, we perform our experiments on over 70 1D univariate time series data from the University of California
Riverside (UCR) time series archive \cite{Bagnall2017TheAdvances,Dau2018TheArchive}. The UCR archive has sub-data sets that span different types of collection methods of the time series data. For simplicity, we call these sub-data sets as data sets which we know come from the UCR data repository. These collection methods include: sensors, spectrograms, image translations, devices, motions, medical, simulations, traffic, speech, acoustic and audio. We focus on the univariate data and try to show the generation can be performed across these various signal types and refer the reader to the UCR archive for more information on each individual data set
used. 

We look at data sets that are binary classification problems as well
as multiclass classification where the most classes taken into consideration
are five. From here, we take a similar approach to Smith et al. in
which we break those data sets into three different categories of
training data size to better compare with the performance of TSGAN
as well as show the ability for uTSGAN to generate time series in
a few shot manner. Our split is 0-499 signals as small training data
sets, 500-1000 signals as medium training data sets, and anything
greater than 1000 as large training data sets. This break down for
training data size is very different than what is typically expected
for training data on image or audio generation tasks \cite{Tsai2015BigSurvey,Chen2014BigPerspectives}. With that in mind, these data sets would all
be small data for image tasks for GANs and one could argue showing
any generations with this time series data would be considered few
shot. 

\subsection{Training Process }

Again, uTSGAN is trained similarly to the original TSGAN. We start
by turning our real time series data into spectrogram images using
a short-time Fourier transform (STFT) \cite{Griffin1984SignalTransform}.
These spectrogram images are used for the real data, $x$, in comparison
for uTSGAN's 2D network, $WGAN_{x}$. We carry no complex information
into the network, rather we are using the RGB values of the spectrogram
image, i.e. a matrix of {[}0-255{]} pixel intensities in each channel.
Since these images are the power spectral density of the signal, we
do realize they carry some spectral contents to them in their shape
and patterns; this is what we feel helps enhance uTSGAN's ability
to generate realistic time series signals. Conditioning $WGAN_{y}$
with these generated spectrograms from $WGAN_{x}$ also helps in the
generation of time series data seeing that relevant spectral information
of the signal showing time frequency characteristics is present in
the condition. $WGAN_{x}$'s discriminator, $D_{x},$ is trained for
some ratio versus the generator, $G$, while at the same time $WGAN_{y}$'s
discriminator, $D_{y}$, and generator, $F$, are trained under the
same ratio. The only caveat to this is the output of $G$ is fed into
$D_{y}$ during the training. We optimize our loss function using
Adam \cite{Kingma2014Adam:Optimization} and our spectrograms are generated and plotted using Librosa
\cite{McFee2015Librosa:Python} and then
saved off as RGB images in JPG format. 

\subsection{Metrics}

We use an adjusted 1D Fréchet Inception Score (FID) \cite{Heusel2017GansEquilibrium}for our metric
to verify the quality of the generated time series data \cite{Hartmann2018EEG-GAN:Signals,Donahue2018AdversarialSynthesis,Engel2019GansS}. The FID score is a way of measuring
the distance between a trained deep network's features learned from
real and synthetic data. These features are then fit into Gaussian
distributions and the Frechet distance (or 2-Wasserstein distance)
is used to measure their similarity. This metric has been shown to
correlate well with images, specifically the quality of an image and
not just how well the classification of the synthetic data is with
a trained classifier (like with the Inception score) \cite{Heusel2017GansEquilibrium}. One issue with this score in the 1D domain is there is no
baseline standard classifier trained on a vast data set to pull these
features from like in the image domain with the Inception \cite{Szegedy2017Inception-v4Learning} classifier
and Imagenet data set \cite{Deng2009ImageNet:Database}. Unlike images,
which share natural inherent features for all natural images, time
series data does not share similarities between collection modalities.
This also makes sense since time series data is so different per application
and the inherent features learned in a ECG signal is different than
that learned in the stock market. So to still be able to use the FID
score to measure our realistic quality of generated data, we create
a unique classifier on each data set and pull features from that classifier.
This of course is slightly skewed in calculation since it would be
ideal to use a classifier with classification accuracy and the high
accuracy on the data set being used this is not always the case, seeing
many of these data sets have top classification rates of 50-70\% with
any classifier. With that, we use a FCN for our time series classification
network to pull features citing it's ability to generalize and perform
well over the entirety of the UCR archive \cite{Wang2017TimeBaseline}. Though
there isn't a standard like the Imagenet data used for image FID metrics,
we know FID score will share similar properties, i.e the lower the
FID score the better the generation is and closer to the original
data. For our results we show the average FID score of the generated
data after 25 runs of generation. 

\subsection{Experimental Setup}

For our experiments we work solely in Ubuntu 16.04 and code our approach
in Python and Tensorflow 2.0 \cite{tensorflow2015-whitepaper,MISCchollet2015keras}
We ran our TSGAN for one thousand epochs on each data set. Our batch
size also differs with every data set trained on, this is due to the
varying size of training data available for each data set. For each
of our data sets we generate samples after 250, 500, 750, and 1000
epochs and then calculate the FID score at those specific epochs.
Our hope is to show that with the combined loss function that uTSGAN
converges faster to better generated results than the TSGAN. Our hardware
is optimized for deep learning, using Nvidia Titan X GPU, an Intel
Core i7 processor, 128G RAM, and a 1TB SSD hard drive. 

\section{Results}

With the span of the 70 different data sets we see data that has extremely
different characteristics. Some of the time series is smooth and follow
similar sine wave oscillation, others are rapid in frequency change
and sharp contrast between time steps with random jumps in values.
This is another reason why we look for a spectral representation of
the signal for the conditional input to $WGAN_{y}$ in order to capture
a lot of these differences in the time series universally. 

\subsection{Empirical Qualitative Analysis}

\begin{figure*}
\begin{centering}
\begin{tabular}{|c|c|}
\hline 
uTSGAN & TSGAN\tabularnewline
\hline 
\includegraphics[scale=0.25]{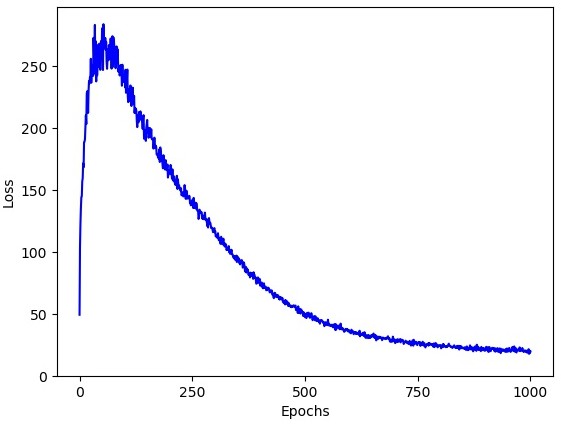} & \includegraphics[scale=0.25]{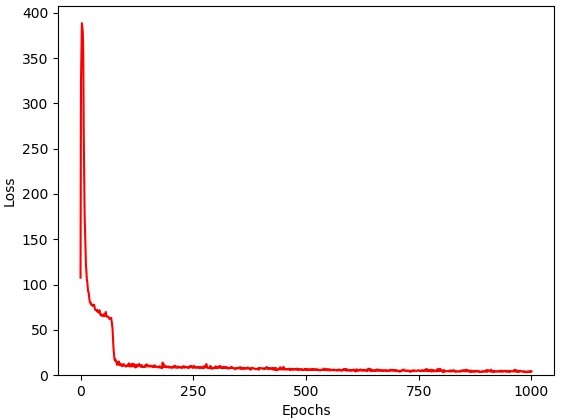} \includegraphics[scale=0.25]{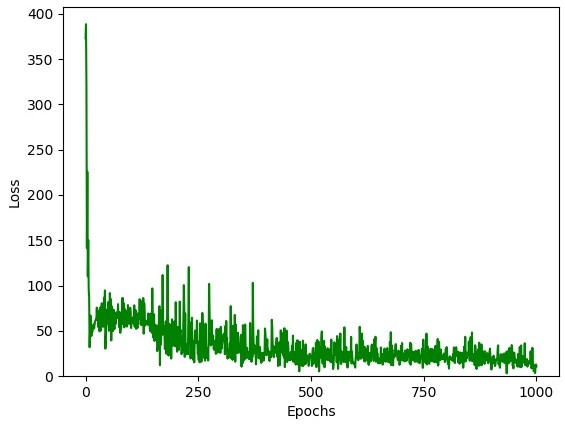}\tabularnewline
\hline
\includegraphics[scale=0.15]{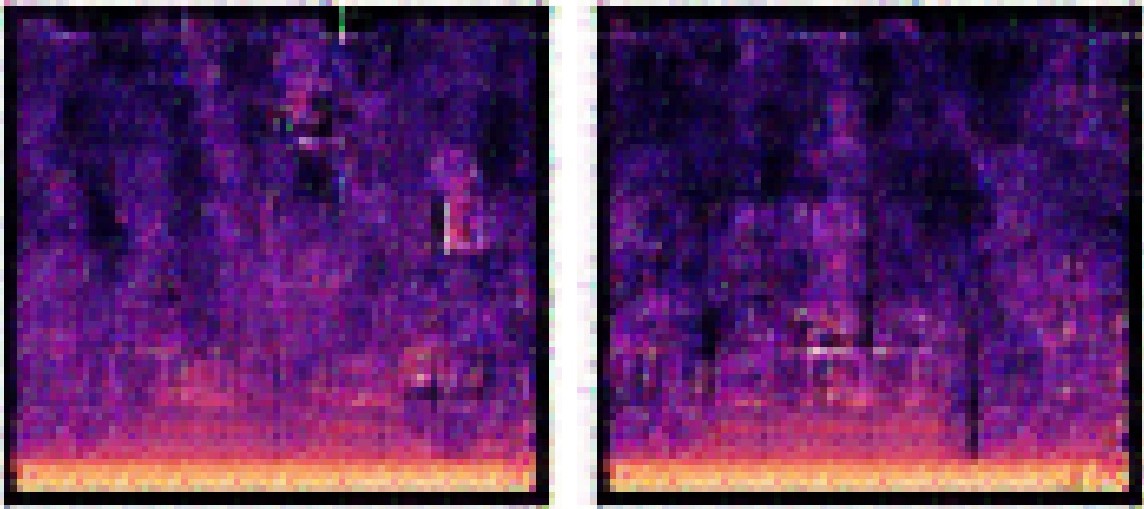} &
\includegraphics[scale=0.159]{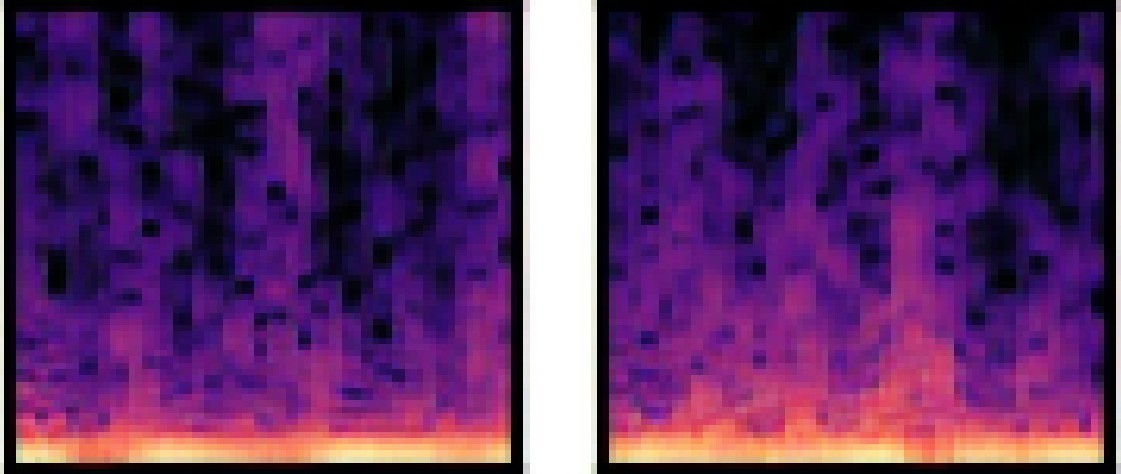}\tabularnewline
\hline
\includegraphics[scale=0.3]{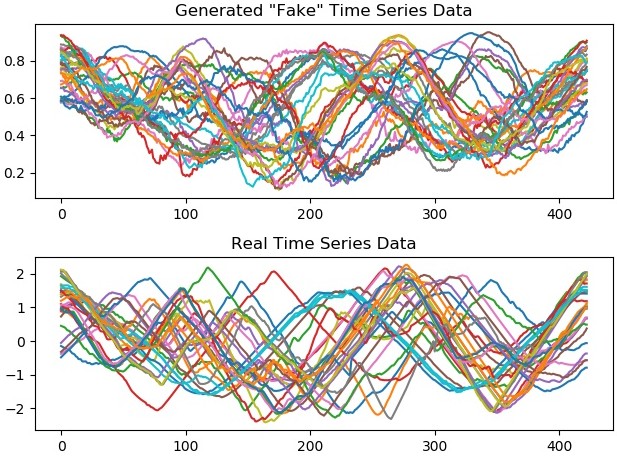} & 
\includegraphics[scale=0.3]{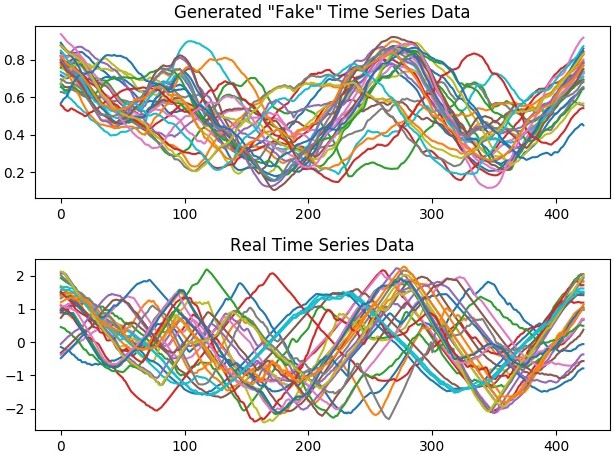}\tabularnewline
\hline

3.36  & 148.30\tabularnewline
\hline 
\end{tabular}
\par\end{centering}
\caption{Yoga, large data set. Left column shows the uTSGAN and the right column is the TSGAN results after 1000 epochs of training.  At the top of the figures we see the loss function graph of each methods. Notice, the TSGAN has two loss graphs, this is due to its two independent networks trained separately. Below these are the generated spectrogram images of class 1, two separate are shown to show generation of different signals spectrum is happening and not just replicating the same spectrogram. Below these are the generated time series signals for class 1 compared to the real signals to see how the generation should look visually.On the bottom of the figure is the FID score of the generated time series, again the lower the FID the better the generation resembles real data. 
\label{figure1:yoga_uTSGANvsTSGAN}}
\label{yoga_loss_plots}
\end{figure*}

\begin{figure*}
\begin{centering}
\begin{tabular}{|c|c|}
\hline 
uTSGAN & TSGAN\tabularnewline
\hline 
\includegraphics[scale=0.25]{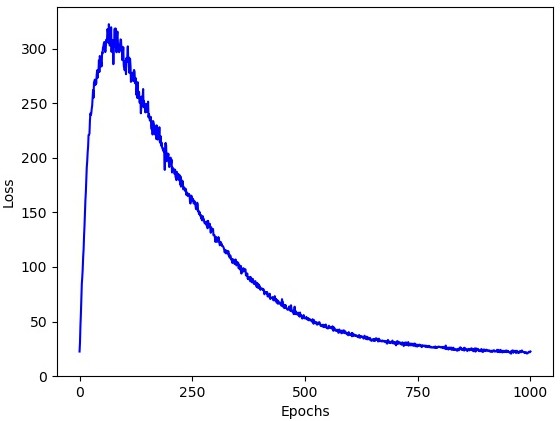} & \includegraphics[scale=0.25]{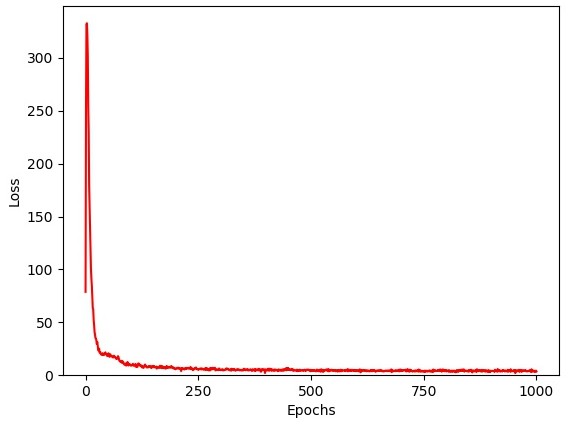} \includegraphics[scale=0.25]{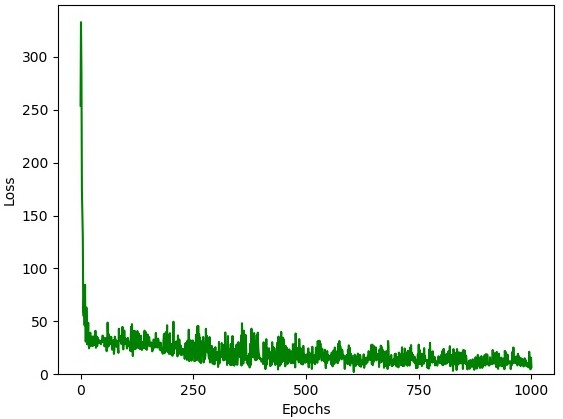}\tabularnewline
\hline
\includegraphics[scale=0.15]{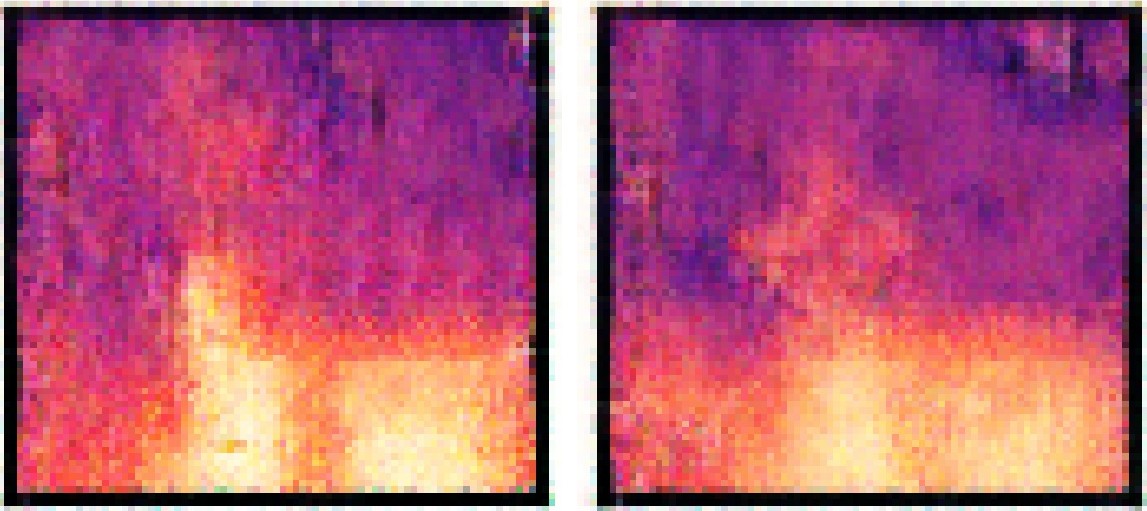} &
\includegraphics[scale=0.159]{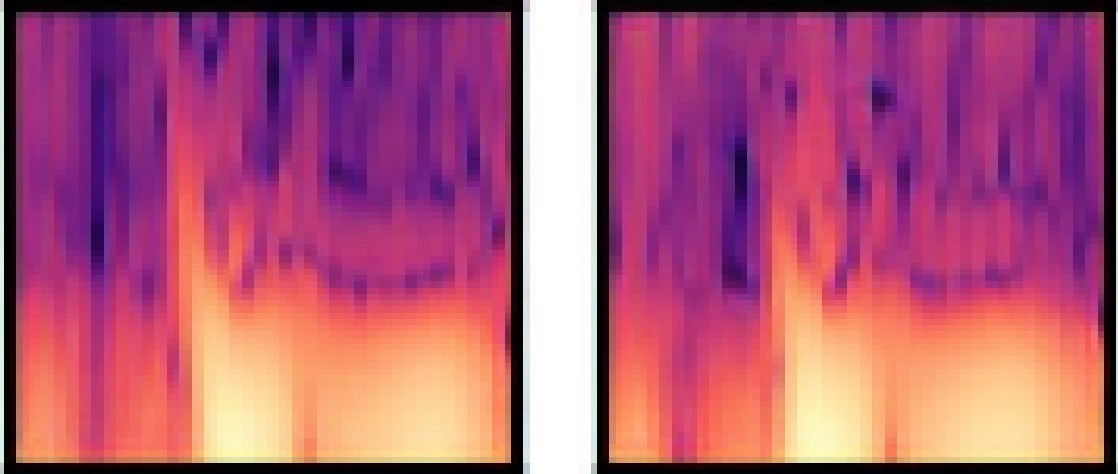}\tabularnewline
\hline
\includegraphics[scale=0.3]{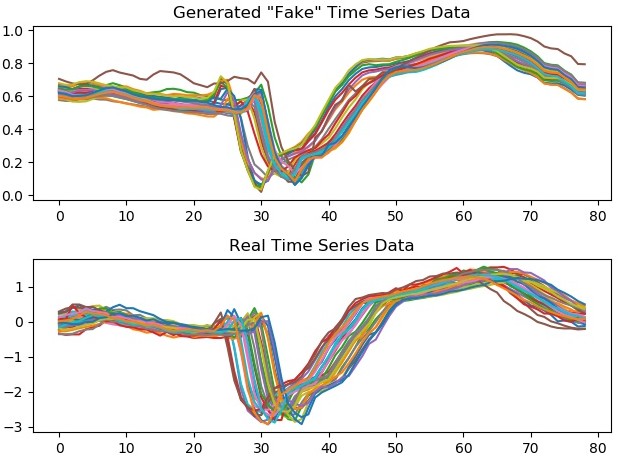} & 
\includegraphics[scale=0.3]{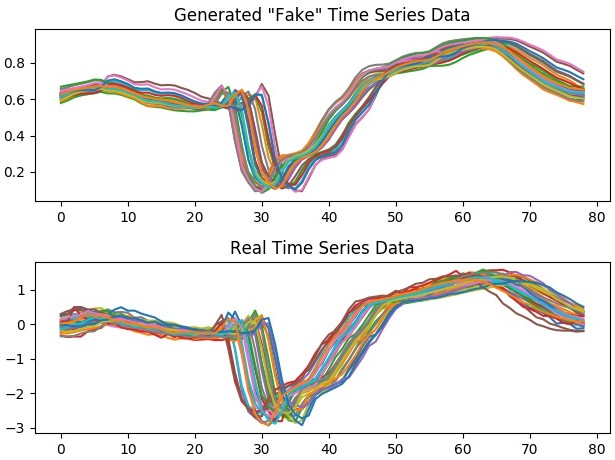}\tabularnewline
\hline
9.25 &  12.23\tabularnewline
\hline 
\end{tabular}
\par\end{centering}
\caption{TwoLeadECG, large data set. Left column shows the uTSGAN and the right column is the TSGAN results after 1000 epochs of training.  At the top of the figures we see the loss function graph of each methods. Notice, the TSGAN has two loss graphs, this is due to its two independent networks trained separately. Below these are the generated spectrogram images of class 1, two separate are shown to show generation of different signals spectrum is happening and not just replicating the same spectrogram. Below these are the generated time series signals for class 1 compared to the real signals to see how the generation should look visually.On the bottom of the figure is the FID score of the generated time series, again the lower the FID the better the generation resembles real data. \label{figure2:two_uTSGANvsTSGAN}}
\label{two_loss_plots}
\end{figure*}

\begin{figure*}
\begin{centering}
\begin{tabular}{|c|c|}
\hline 
uTSGAN & TSGAN\tabularnewline
\hline 
\includegraphics[scale=0.25]{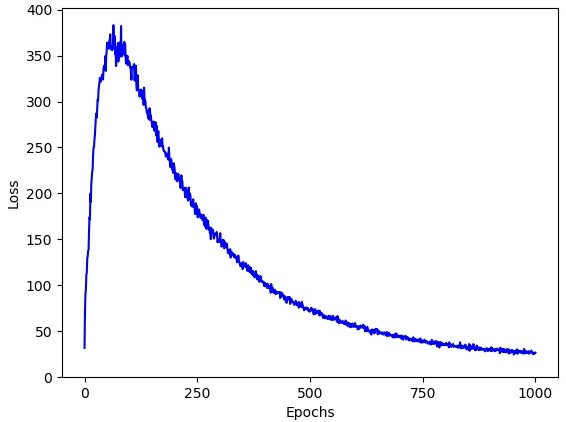} & \includegraphics[scale=0.25]{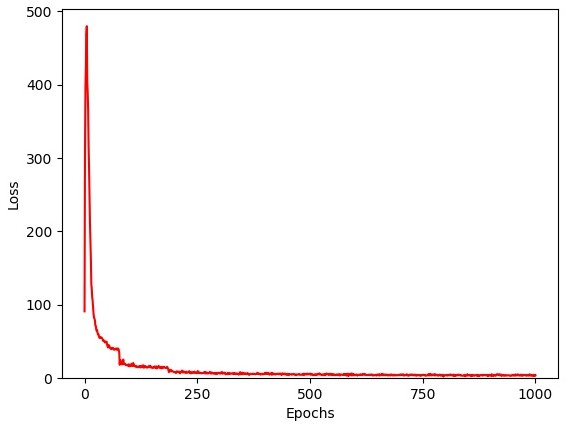} \includegraphics[scale=0.25]{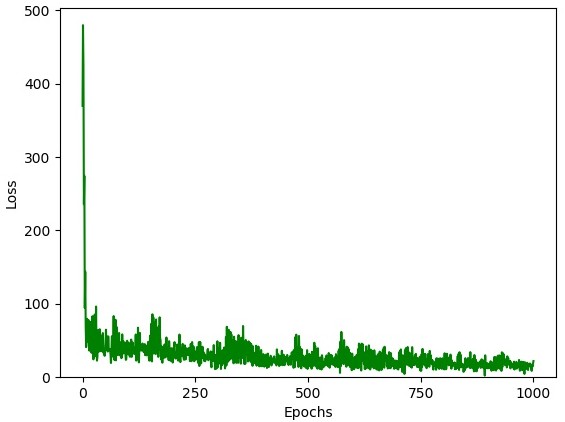}\tabularnewline
\hline
\includegraphics[scale=0.15]{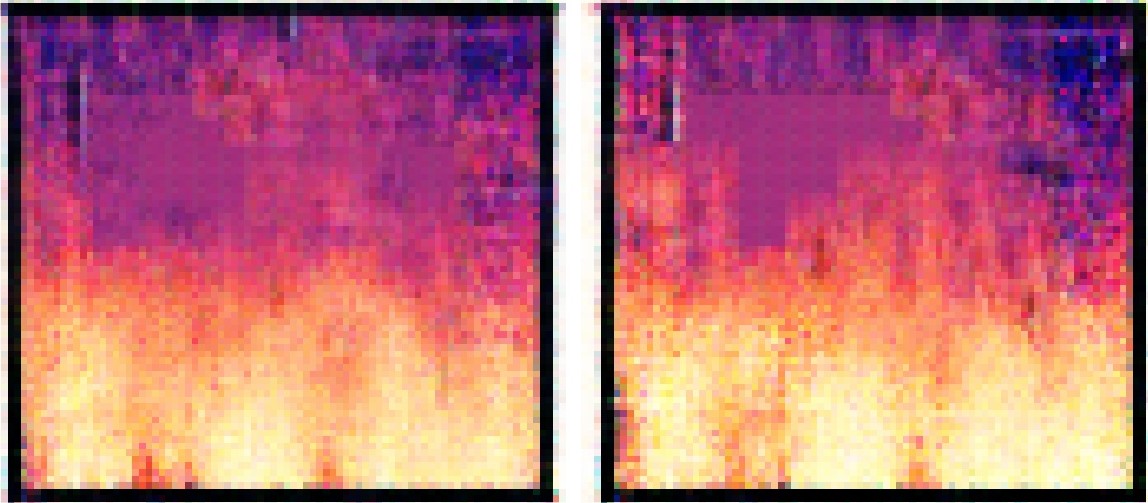} &
\includegraphics[scale=0.159]{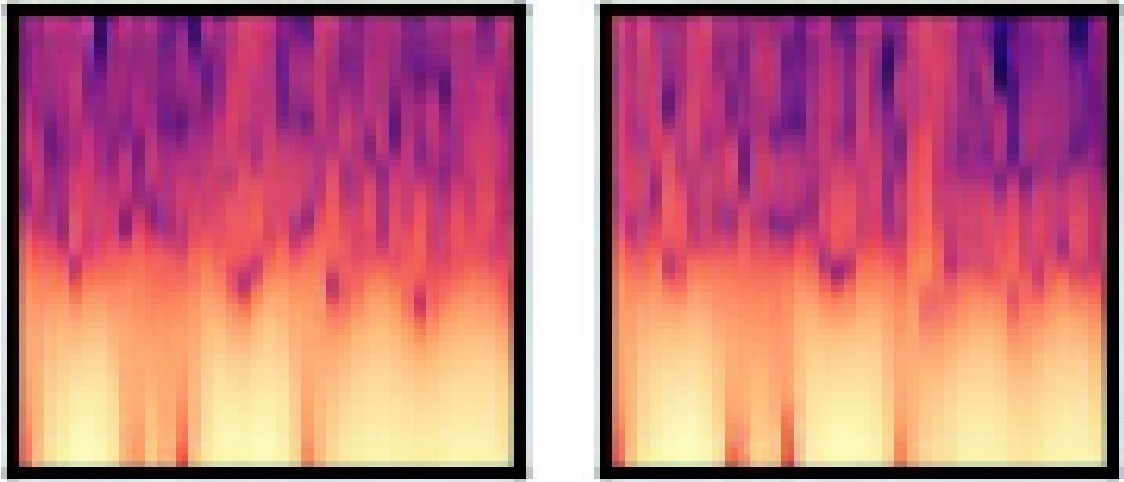}\tabularnewline
\hline
\includegraphics[scale=0.3]{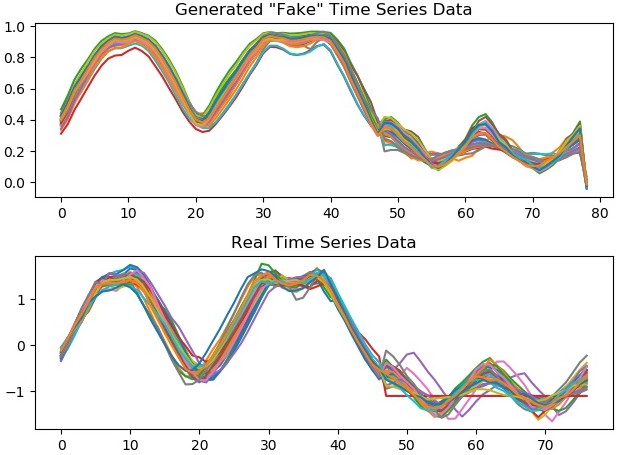} & 
\includegraphics[scale=0.3]{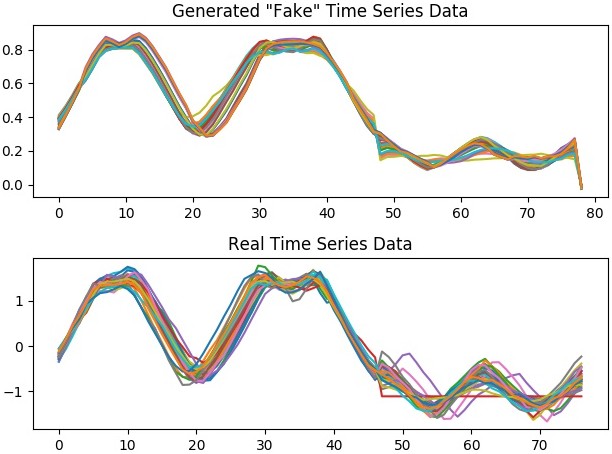}\tabularnewline
\hline
72.99 &  174.70\tabularnewline
\hline 
\end{tabular}
\par\end{centering}
\caption{MiddlePhalanxOutlineCorrect, medium data set. Left column shows the uTSGAN and the right column is the TSGAN results after 1000 epochs of training.  At the top of the figures we see the loss function graph of each methods. Notice, the TSGAN has two loss graphs, this is due to its two independent networks trained separately. Below these are the generated spectrogram images of class 0, two separate are shown to show generation of different signals spectrum is happening and not just replicating the same spectrogram. Below these are the generated time series signals for class 0 compared to the real signals to see how the generation should look visually.On the bottom of the figure is the FID score of the generated time series, again the lower the FID the better the generation resembles real data. \label{figure3:middle_uTSGANvsTSGAN}}
\label{middle_loss_plots}
\end{figure*}

\begin{figure*}
\begin{centering}
\begin{tabular}{|c|c|}
\hline 
uTSGAN & TSGAN\tabularnewline
\hline 
\includegraphics[scale=0.25]{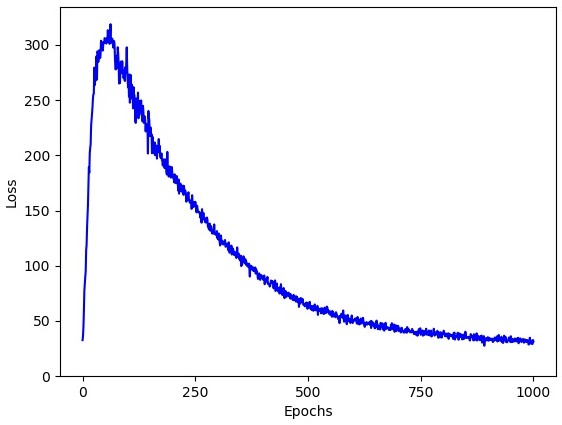} & \includegraphics[scale=0.25]{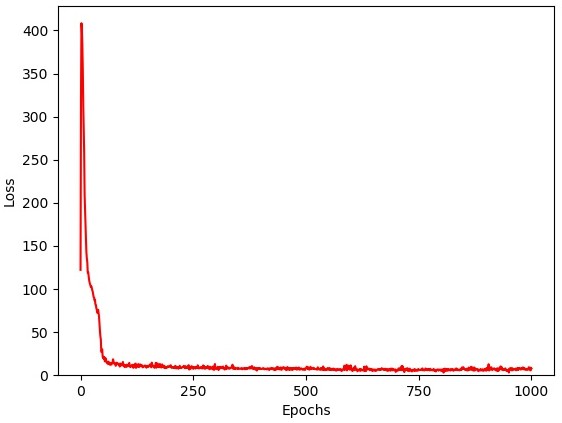} \includegraphics[scale=0.25]{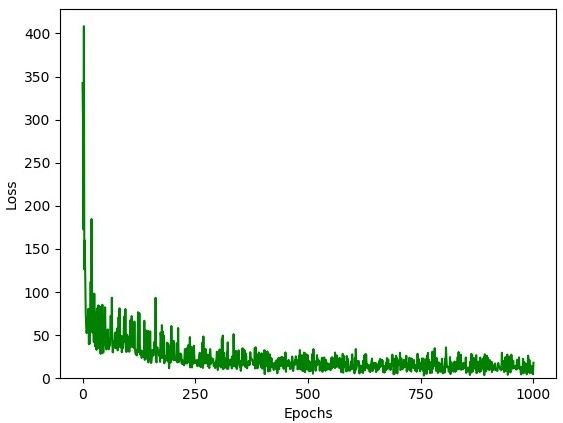}\tabularnewline
\hline
\includegraphics[scale=0.15]{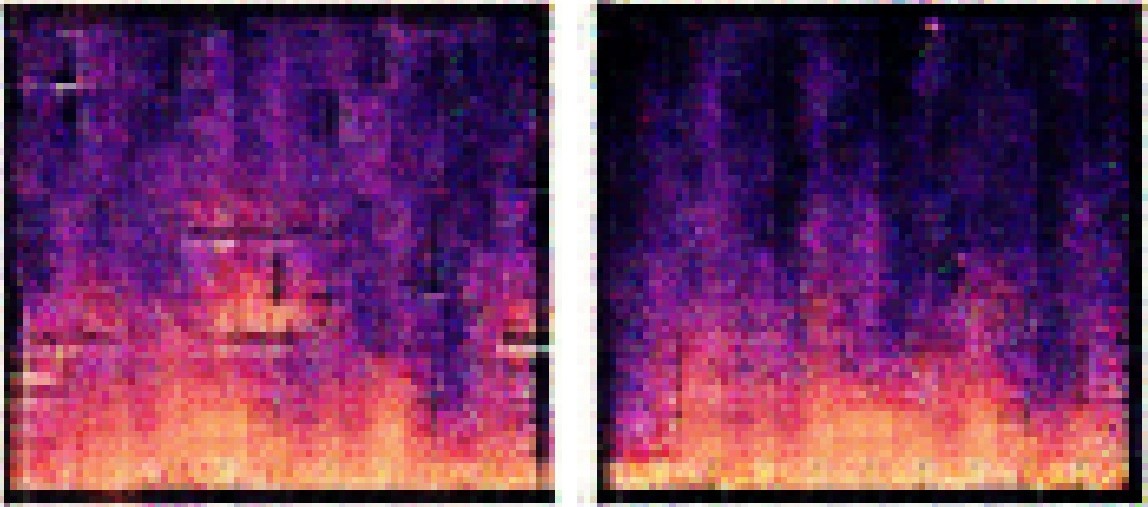} &
\includegraphics[scale=0.159]{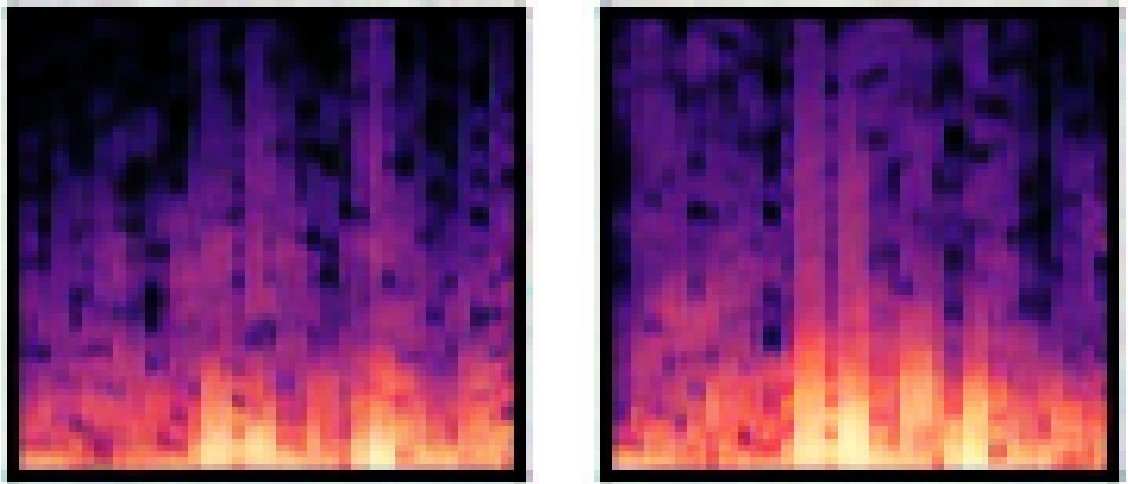}\tabularnewline
\hline
\includegraphics[scale=0.3]{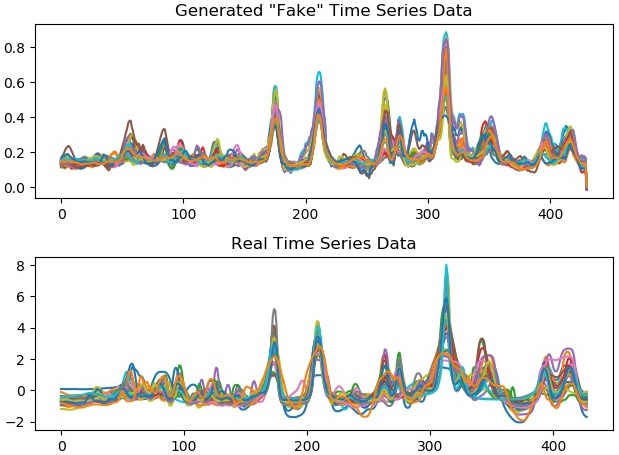} & 
\includegraphics[scale=0.3]{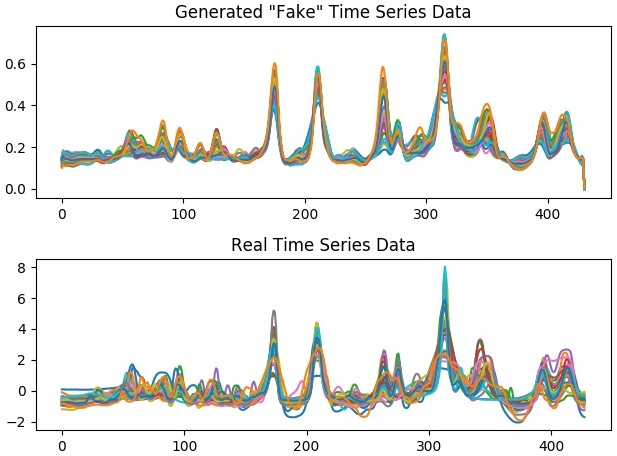}\tabularnewline
\hline

26.30 & 89.26\tabularnewline
\hline 
\end{tabular}
\par\end{centering}
\caption{Ham, small data set. Left column shows the uTSGAN and the right column is the TSGAN results after 1000 epochs of training.  At the top of the figures we see the loss function graph of each methods. Notice, the TSGAN has two loss graphs, this is due to its two independent networks trained separately. Below these are the generated spectrogram images of class 1, two separate are shown to show generation of different signals spectrum is happening and not just replicating the same spectrogram. Below these are the generated time series signals for class 1 compared to the real signals to see how the generation should look visually.On the bottom of the figure is the FID score of the generated time series, again the lower the FID the better the generation resembles real data. \label{figure4:ham_uTSGANvsTSGAN}}
\label{ham_loss_plots}
\end{figure*}

\begin{figure*}
\begin{centering}
\begin{tabular}{|c|c|}
\hline 
uTSGAN & TSGAN\tabularnewline
\hline 
\includegraphics[scale=0.25]{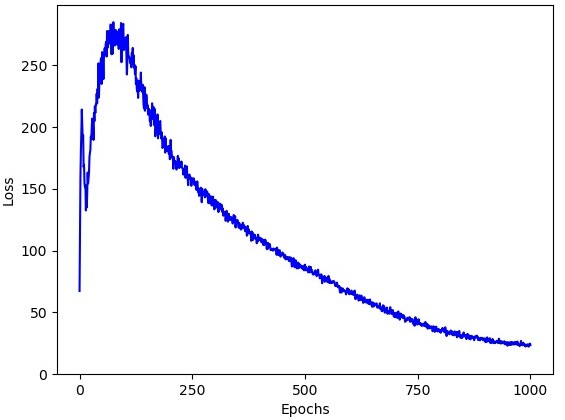} & \includegraphics[scale=0.25]{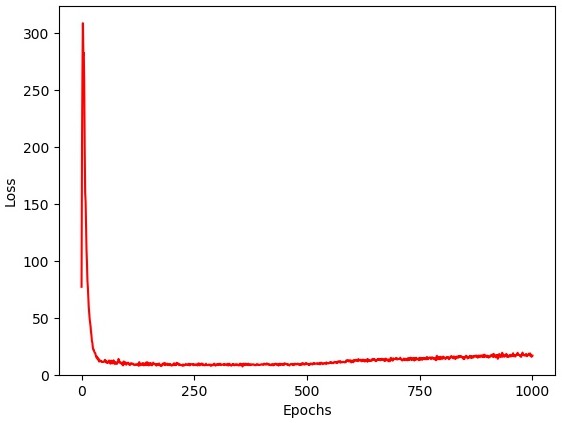} \includegraphics[scale=0.25]{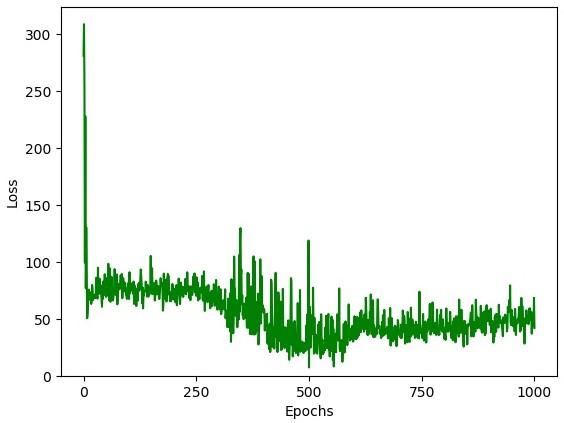}\tabularnewline
\hline
\includegraphics[scale=0.15]{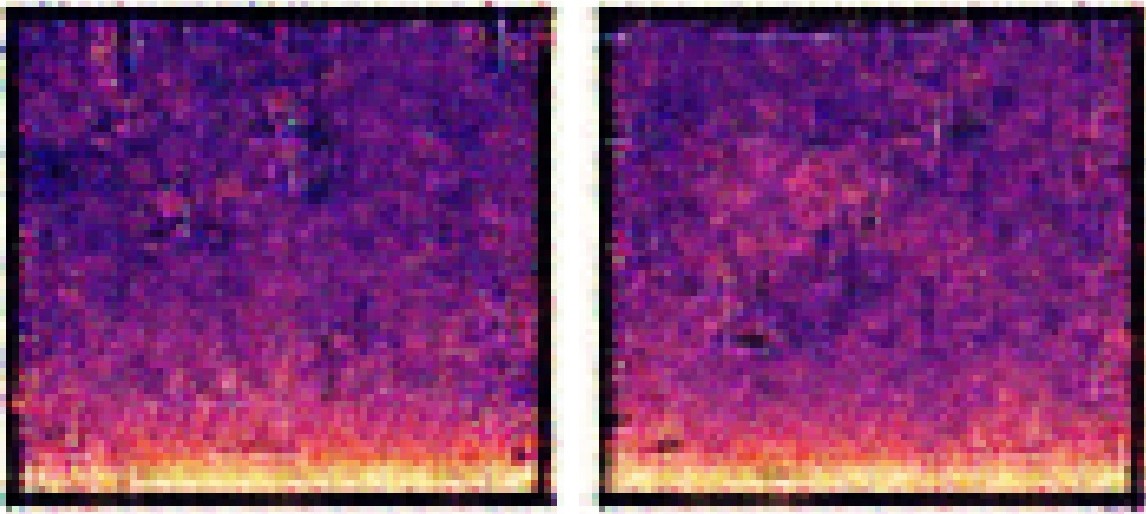} &
\includegraphics[scale=0.159]{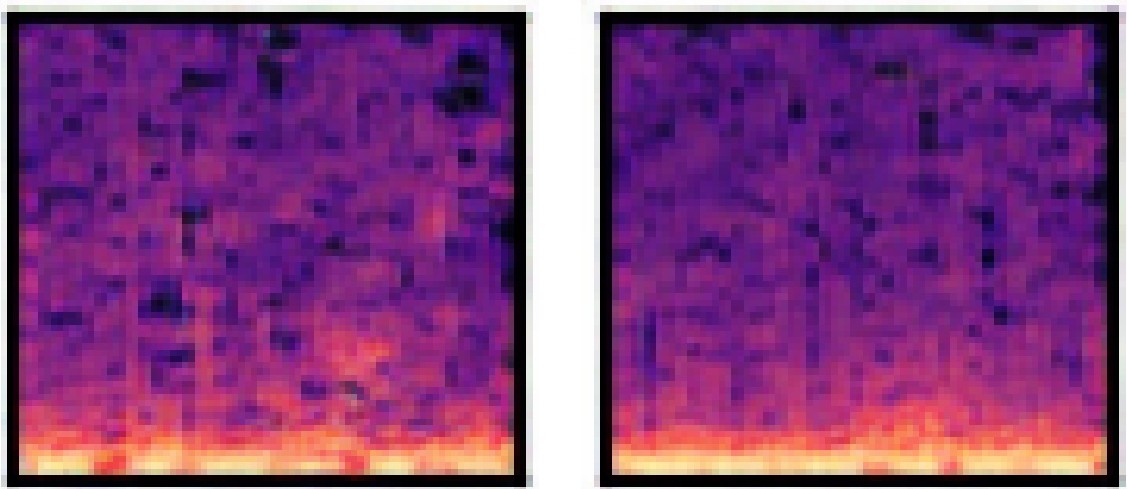}\tabularnewline
\hline
\includegraphics[scale=0.3]{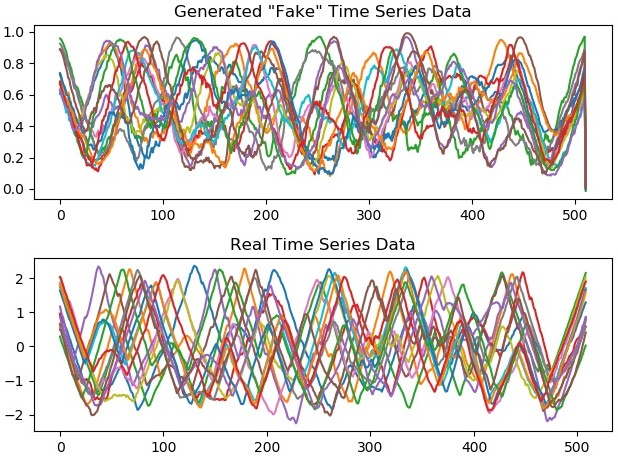} & 
\includegraphics[scale=0.3]{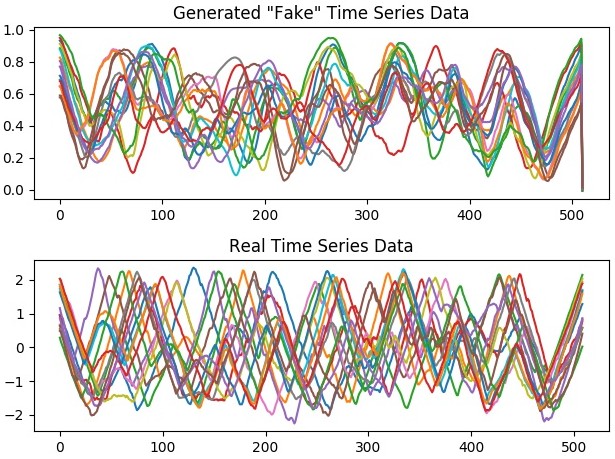}\tabularnewline
\hline

14.61 & 26.20\tabularnewline
\hline 
\end{tabular}
\par\end{centering}
\caption{BeetleFly, small data set. Left column shows the uTSGAN and the right column is the TSGAN results after 1000 epochs of training.  At the top of the figures we see the loss function graph of each methods. Notice, the TSGAN has two loss graphs, this is due to its two independent networks trained separately. Below these are the generated spectrogram images of class 1, two separate are shown to show generation of different signals spectrum is happening and not just replicating the same spectrogram. Below these are the generated time series signals for class 1 compared to the real signals to see how the generation should look visually.On the bottom of the figure is the FID score of the generated time series, again the lower the FID the better the generation resembles real data. \label{figure5:beetle_uTSGANvsTSGAN}}
\label{beetle_loss_plots}
\end{figure*}

\begin{figure*}
\begin{centering}
\begin{tabular}{|c|c|}
\hline 
uTSGAN & TSGAN\tabularnewline
\hline 
\includegraphics[scale=0.25]{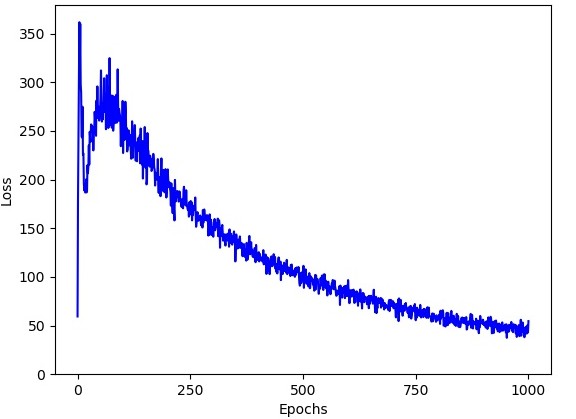} & \includegraphics[scale=0.25]{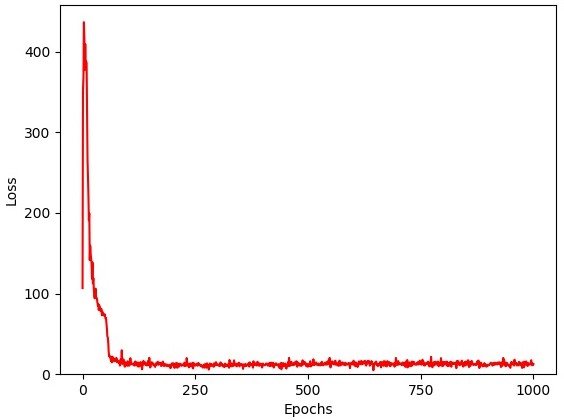} \includegraphics[scale=0.25]{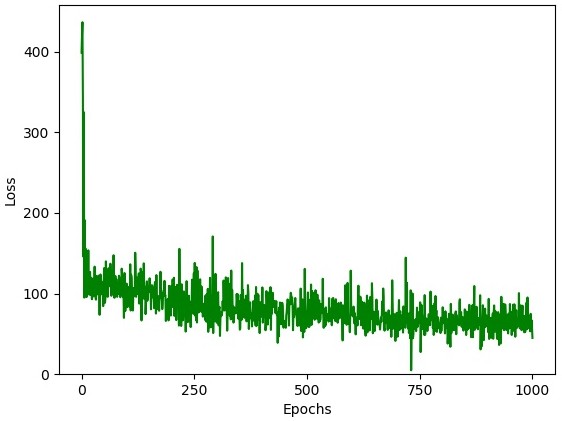}\tabularnewline
\hline
\includegraphics[scale=0.15]{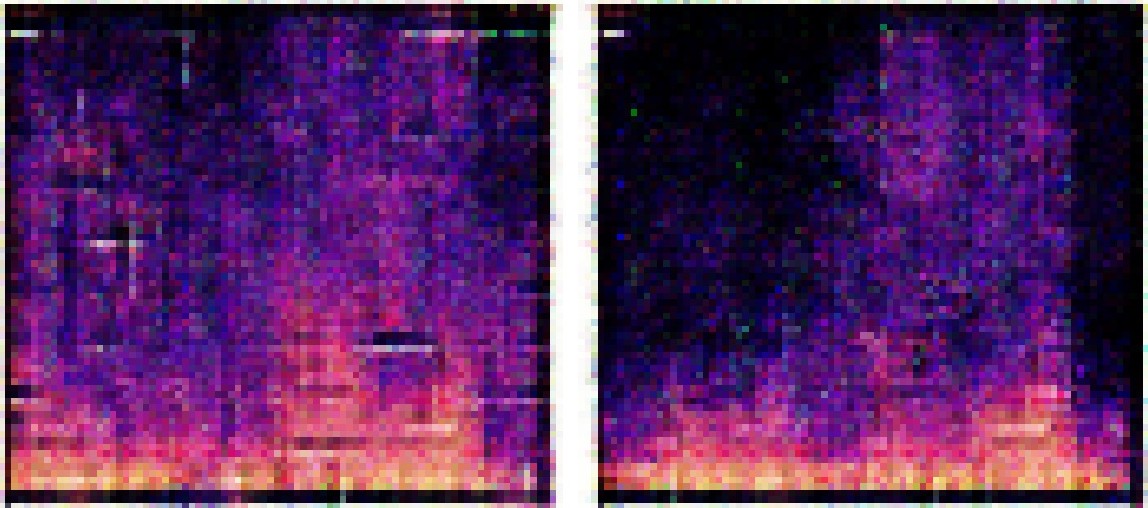} &
\includegraphics[scale=0.159]{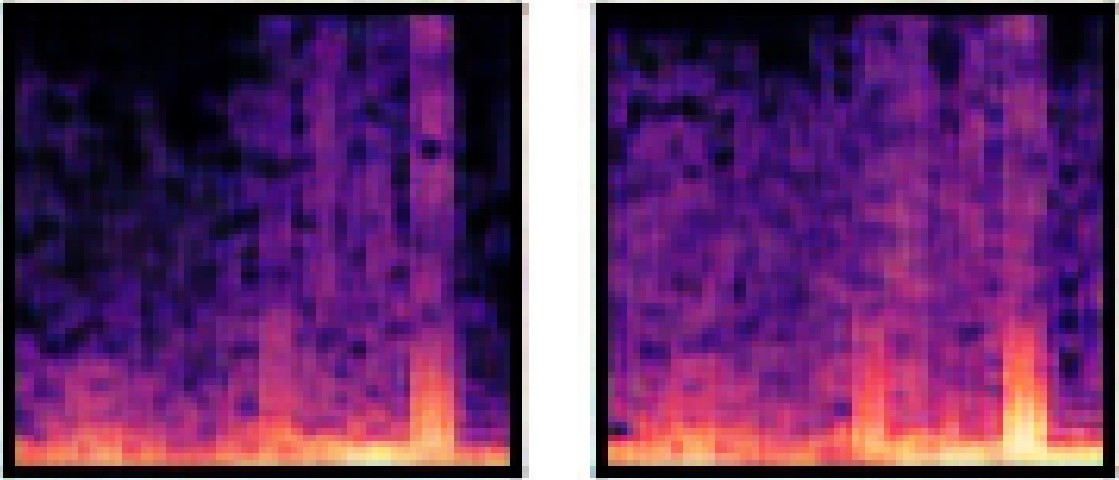}\tabularnewline
\hline
\includegraphics[scale=0.3]{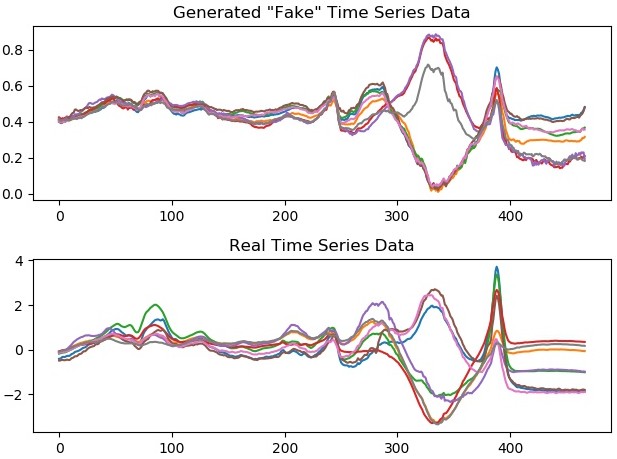} & 
\includegraphics[scale=0.3]{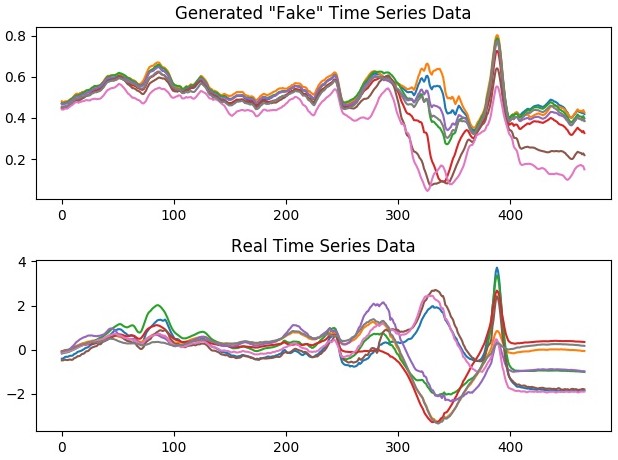}\tabularnewline
\hline

47.26 & 207.10\tabularnewline
\hline 
\end{tabular}
\par\end{centering}
\caption{Beef, small data set. Left column shows the uTSGAN and the right column is the TSGAN results after 1000 epochs of training.  At the top of the figures we see the loss function graph of each methods. Notice, the TSGAN has two loss graphs, this is due to its two independent networks trained separately. Below these are the generated spectrogram images of class 1, two separate are shown to show generation of different signals spectrum is happening and not just replicating the same spectrogram. Below these are the generated time series signals for class 1 compared to the real signals to see how the generation should look visually. On the bottom of the figure is the FID score of the generated time series, again the lower the FID the better the generation resembles real data. \label{figure6:beef_uTSGANvsTSGAN}}
\label{beef_loss_plots}
\end{figure*}

To begin, we look at the unified loss function of uTSGAN plotted versus
number of epochs and in comparison both independent loss functions
for TSGAN. Some of the data sets loss function can be seen in Fig \ref{yoga_loss_plots} - Fig \ref{beef_loss_plots} which all show the comparison of different data sets. Here we
have two large grouped data sets, one medium, and three small. Diving
in further we can look at similarities between all these images as
well as the generation of the time series. To begin, we see that the
loss function for uTSGAN follows similar paths for all data sets shown.
We believe this behavior is due to the invariant penalty of the network
on itself by having to generate good spectrogram images in order to
generate good time series data. When the time series data isn't generated
correctly, the network has to relearn the spectrogram generation in
order to better suit the system. This can be seen as a better dependency relationship which requires the network
to learn more efficiently and create better time series data. More
on this subject further down this section. 

We also see in Fig \ref{yoga_loss_plots} - Fig \ref{beef_loss_plots} that TSGAN's original loss for $WGAN_{x}$
converges extremely quickly while the loss for $WGAN_{y}$ does not.
We believe this is because of the ability of WGANs to generate synthetic
images extremely well to begin with and is their core function in
the community. On top of that, spectrogram images do not share the
same features or temporal conditions that natural images share which
makes their generation an easier task than generating faces or animals
that don't look like Picasso paintings. 

Digging deeper into the dependency issue first realized in the original
TSGAN, we look at the loss of $WGAN_{x}$ and try to compare it to
the generated time series from $WGAN_{y}$. Originally, there is no
dependency mathematically related in the losses of the two independent networks in TSGAN, $WGAN_{x}$ and $WGAN_{y}$,
which means that $WGAN_{x}$ can be feeding $WGAN_{y}$ great realistic
spectrograms or horrible ones with no penalty. This is essentially
what is happening with two independent loss functions. As we can see
in Fig \ref{old_tsgan_spectrogramComapre} the original TSGAN's $WGAN_{x}$ is generating
realistic spectrogram images from as little as 100 epochs of training
(this is also seen throughout the data sets trained, Fig \ref{old_tsgan_spectrogramComapre} we show a few). What is not happening though is the same quality
of generation being done by $WGAN_{y}$ for the time series data in
as little epochs as the spectrogram images and even after all training
is done.

This shows that the condition of the spectrogram image to $WGAN_{y}$
in the original TSGAN is really not conditioning the network at all.
What is really happening is a random input matrix is being given to
$WGAN_{y}$ (which in this case is the generated spectrogram image
generated from $WGAN_{x}$ per epoch) and then used for generating
the time series data. This is no different than the original GAN method
of generation by feeding some random vector to the generator every
epoch, only in this case the random vector is in the shape of a random
generated RGB values that are drawn from the space of generated spectrogram
images. With this, it is impressive that TSGAN originally was able
to generate such quality time series data as well as demonstrate few
shot ability. We believe this is due to the power of the architecture
in $WGAN_{y}$ which as described by Smith et al. as a 2D to 1D autoencoder.
In there more information must be given to $WGAN_{y}$ to generate
time series data from the random matrix instead of a random $N\times1$
random input vector. 

\begin{figure*}[h!]
\begin{centering}
\begin{tabular}{|c|c|c|c|}
\hline 
Epoch 250 & Epoch 500 & Epoch 750 & Epoch 1000\tabularnewline
\hline
\includegraphics[scale=0.09]{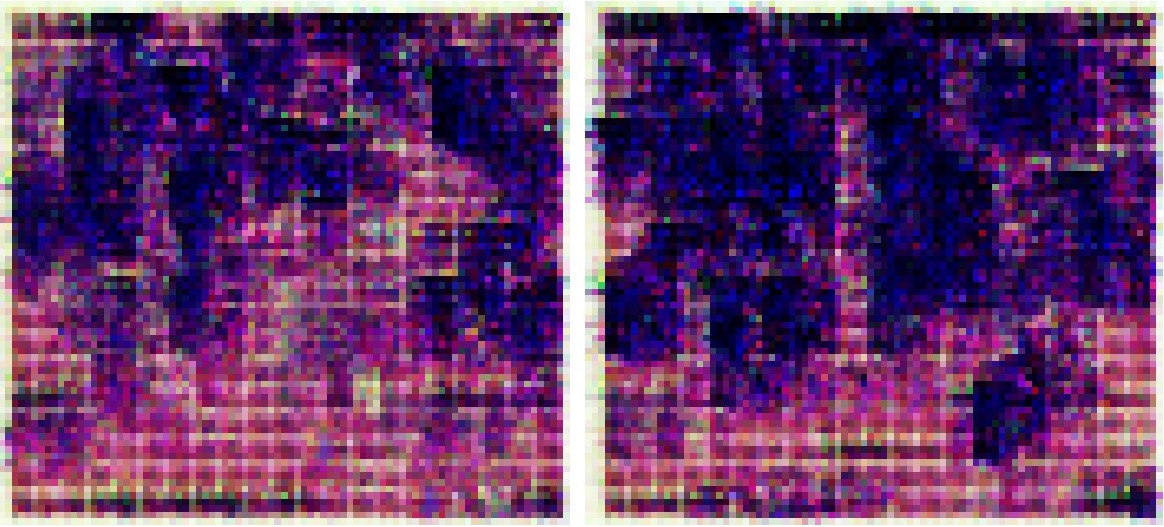} &
\includegraphics[scale=0.09]{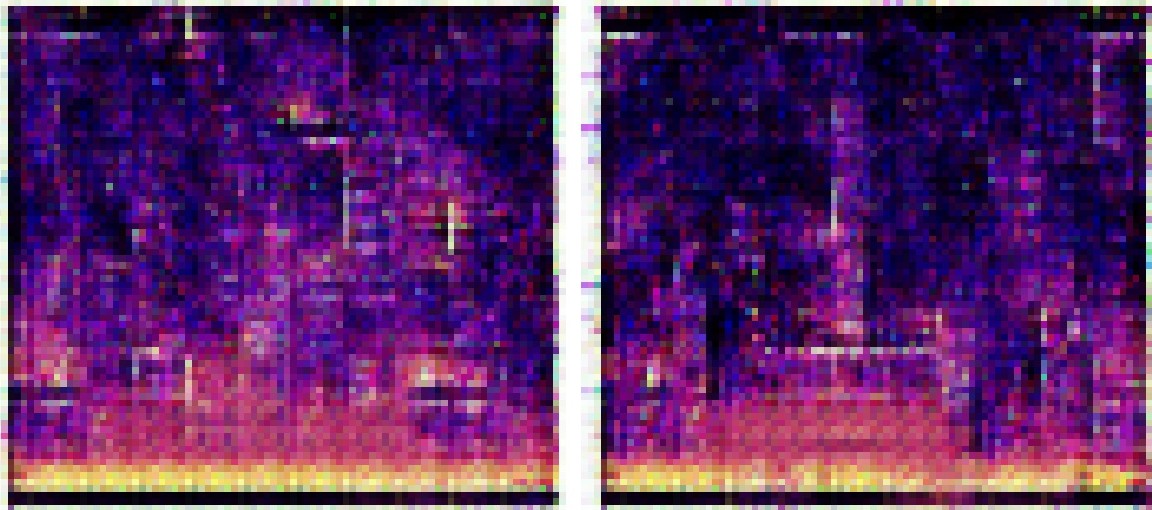} & 
\includegraphics[scale=0.09]{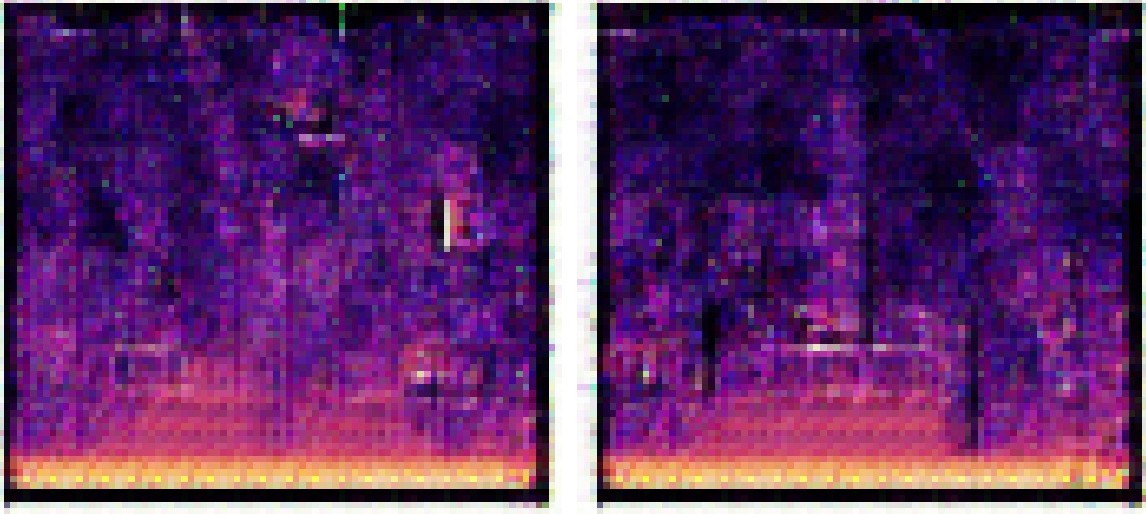} & 
\includegraphics[scale=0.09]{figures/uTsgan_yoga_2D1000.jpg}\tabularnewline

\includegraphics[scale=0.17]{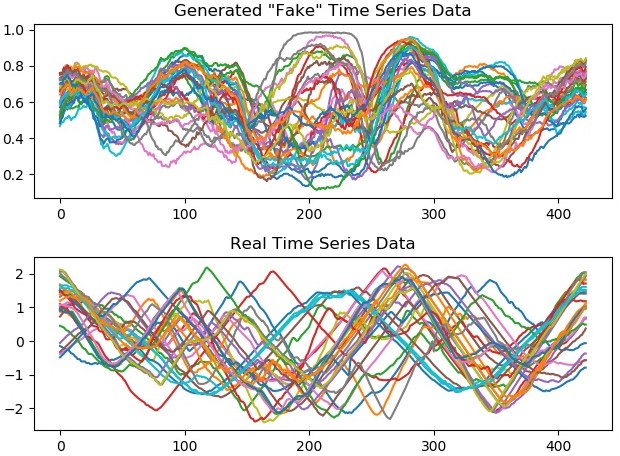} & 
\includegraphics[scale=0.17]{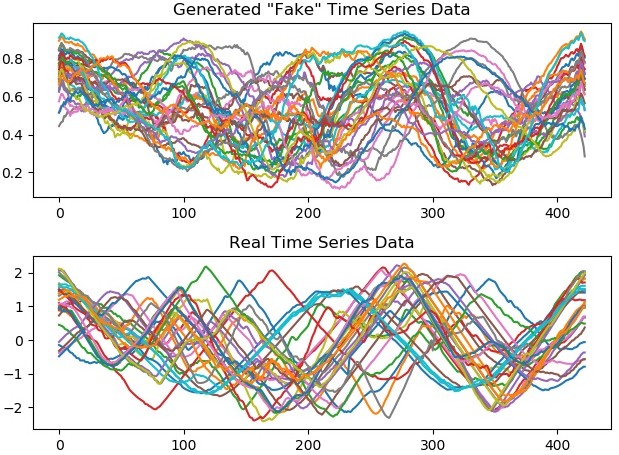} &
\includegraphics[scale=0.17]{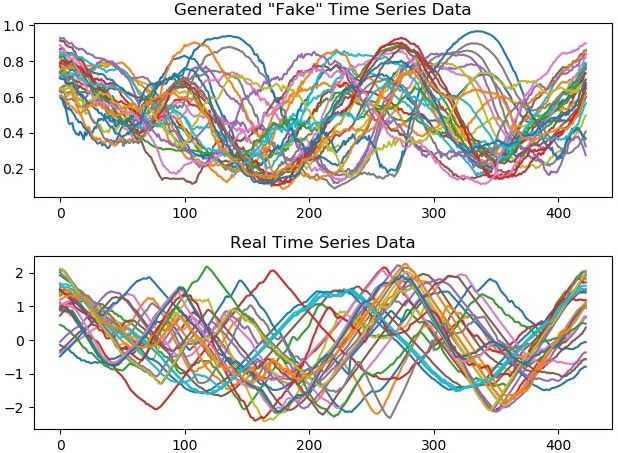} & 
\includegraphics[scale=0.17]{figures/uTsgan_yoga_1000.jpg}\tabularnewline
\hline 
65.94 & 31.06 & 18.04 & 3.36\tabularnewline
\hline 
\end{tabular}
\par\end{centering}
\caption{Yoga class 1, large data set. Examples of spectrogram images and time series pair generated from uTSGAN for associated epochs and the FID score at each epoch. This shows uTSGAN consistently gets better at generation over training. It also depicts the dependency of the two networks, showing as the quality of the spectrogram gets better so does the time series generation. }
\label{yoga_utsgan_epoch_results}
\end{figure*}

\begin{figure*}[h!]
\begin{centering}
\begin{tabular}{|c|c|c|c|}
\hline 
Epoch 250 & Epoch 500 & Epoch 750 & Epoch 1000\tabularnewline
\hline 
\includegraphics[scale=0.09]{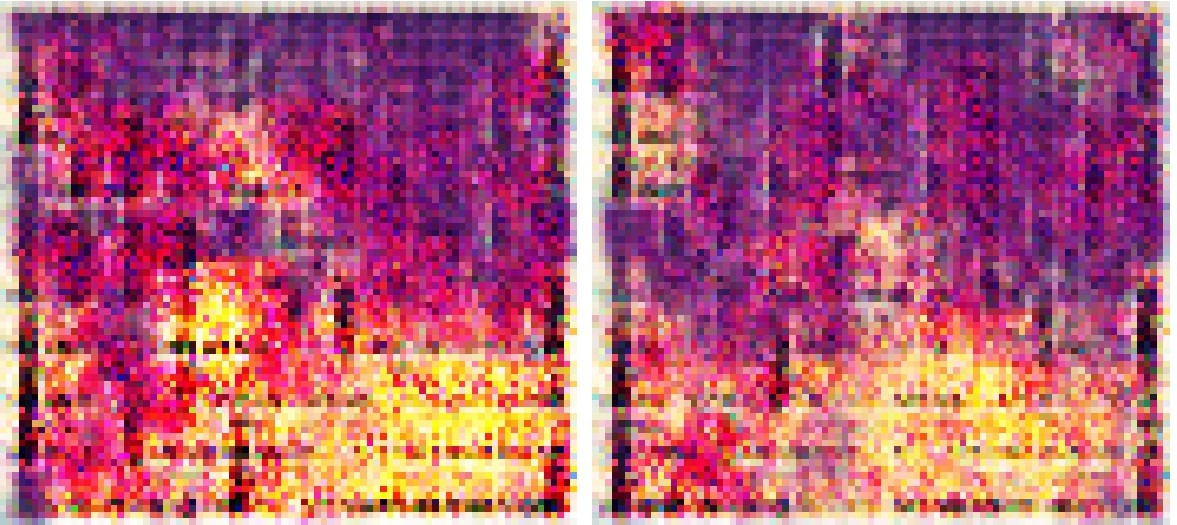} & 
\includegraphics[scale=0.09]{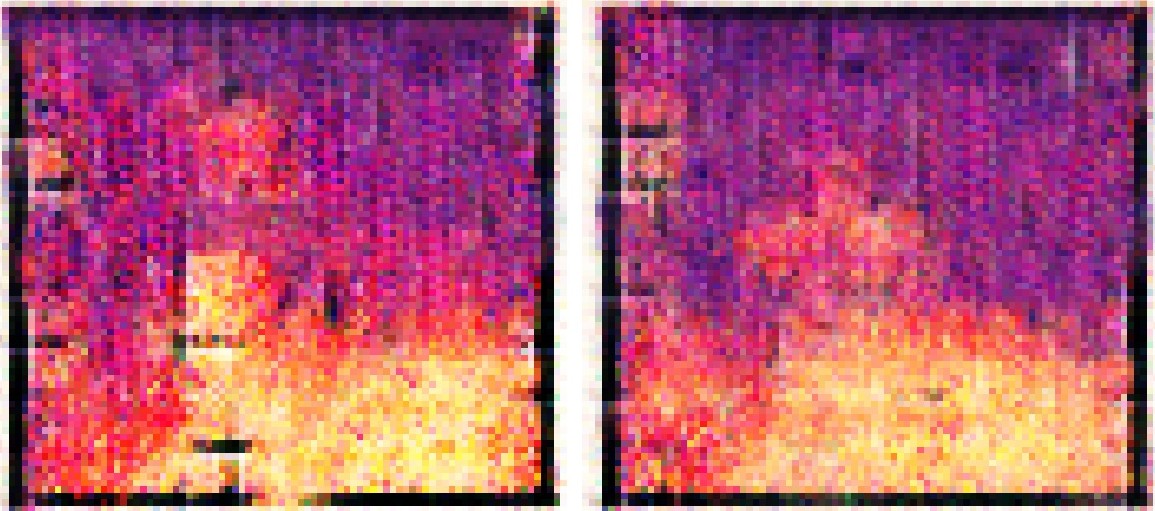} & 
\includegraphics[scale=0.09]{figures/uTsgan_two_2D1000.jpg} &
\includegraphics[scale=0.09]{figures/uTsgan_two_2D1000.jpg}\tabularnewline
 
\includegraphics[scale=0.17]{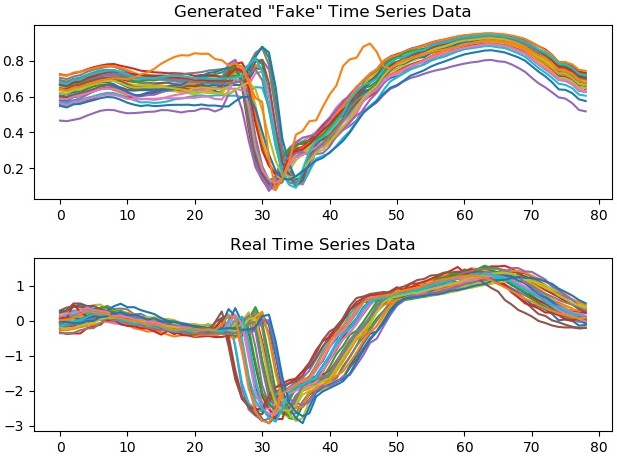} & 
\includegraphics[scale=0.17]{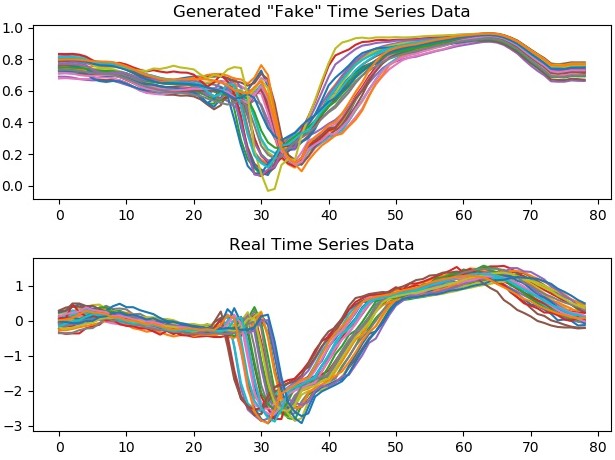} & 
\includegraphics[scale=0.17]{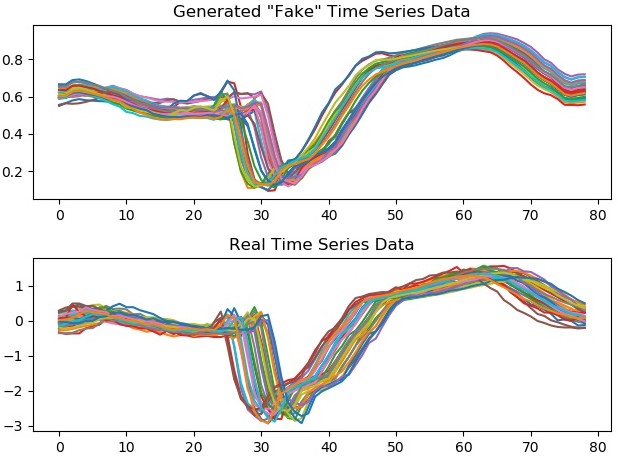} &
\includegraphics[scale=0.17]{figures/uTsgan_two_1000.jpg}\tabularnewline
\hline 
27.24 & 18.02 & 10.30 & 9.25\tabularnewline
\hline 
\end{tabular}
\par\end{centering}
\caption{TwoLeadECG class 1, large data set. Examples of spectrogram images and time series pair generated from uTSGAN for associated epochs and the FID score at each epoch. This shows uTSGAN consistently gets better at generation over training. It also depicts the dependency of the two networks, showing as the quality of the spectrogram gets better so does the time series generation.}
\label{twoleadECG_utsgan_epoch_results}
\end{figure*}

\begin{figure*}[h!]
\begin{centering}
\begin{tabular}{|c|c|c|c|}

\hline 
Epoch 250 & Epoch 500 & Epoch 750 & Epoch 1000\tabularnewline
\hline
\includegraphics[scale=0.09]{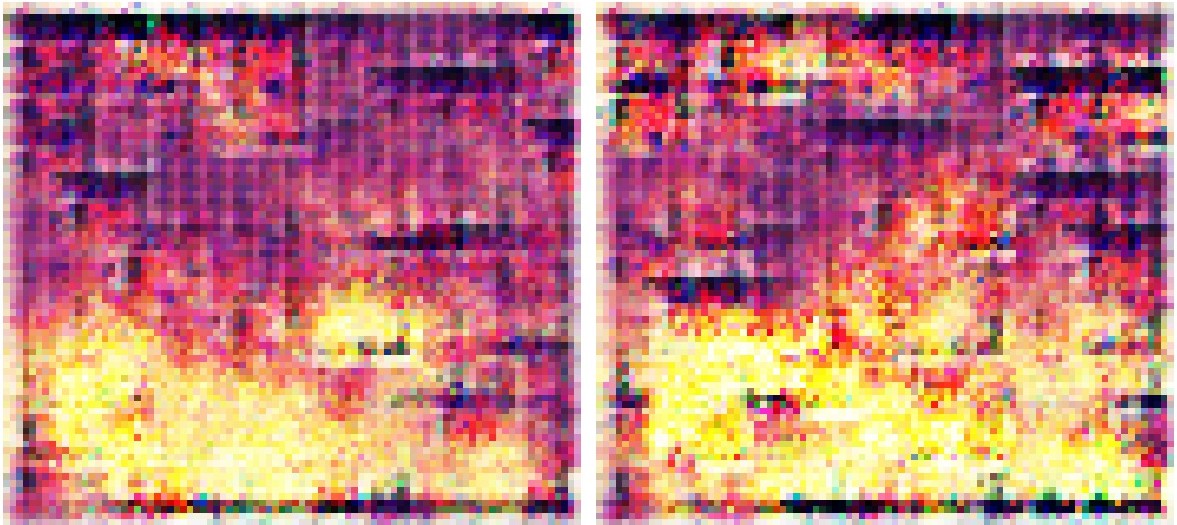} & 
\includegraphics[scale=0.09]{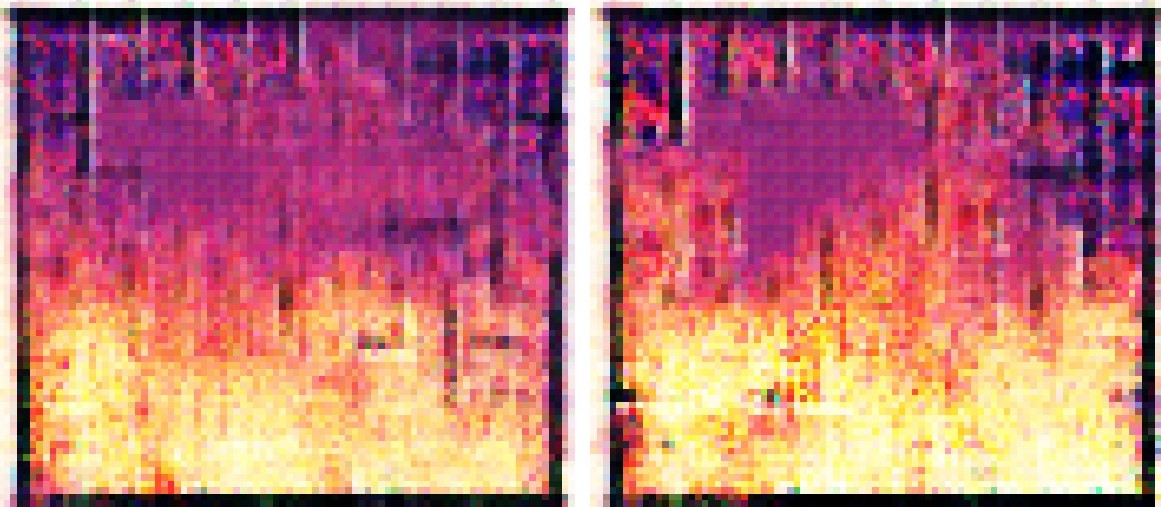} & 
\includegraphics[scale=0.09]{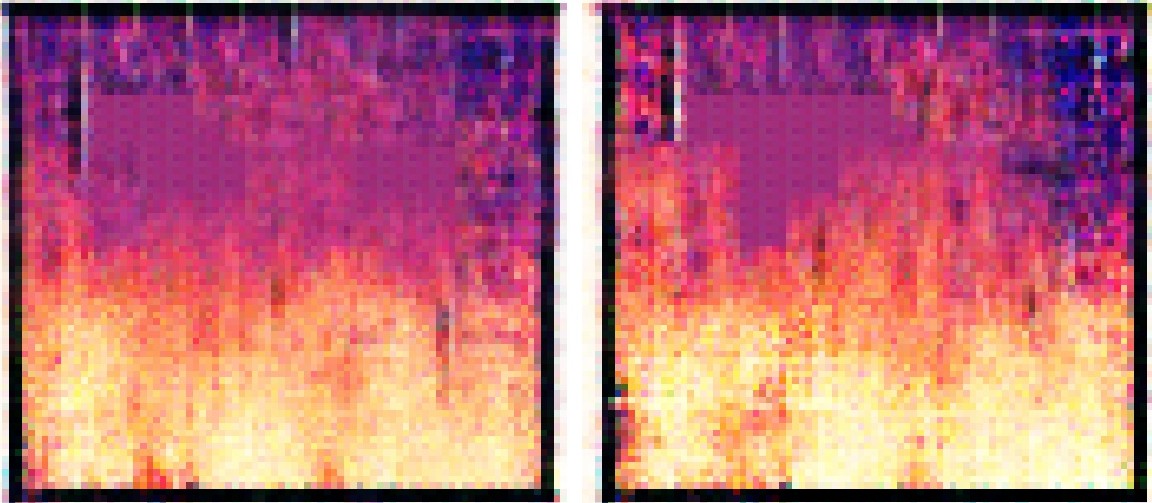} &
\includegraphics[scale=0.09]{figures/uTsgan_mid_2D1000.jpg}\tabularnewline
 
\includegraphics[scale=0.17]{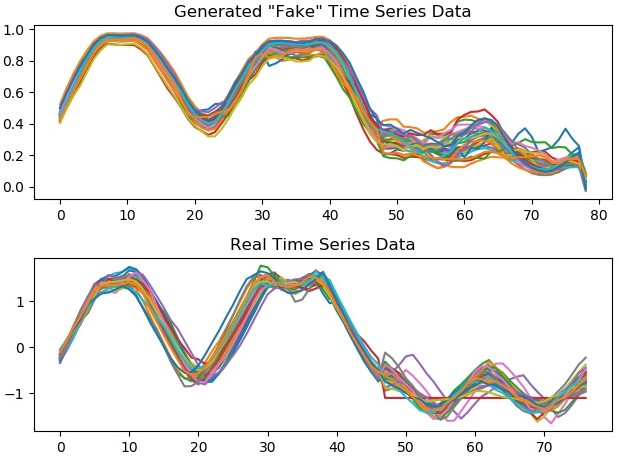} & 
\includegraphics[scale=0.17]{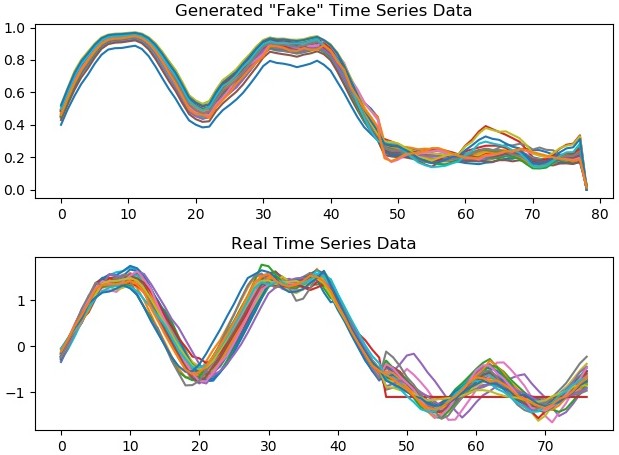} & 
\includegraphics[scale=0.17]{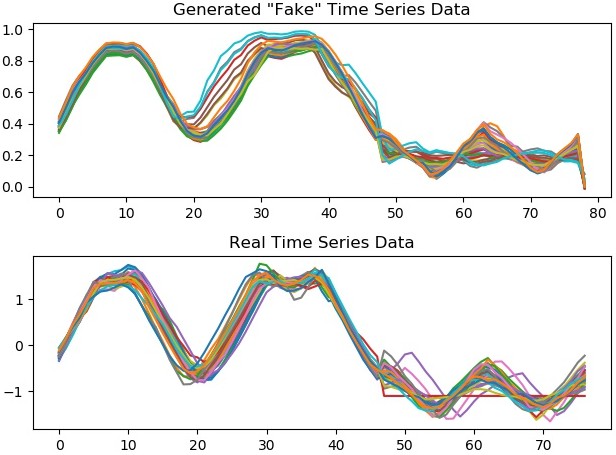} &
\includegraphics[scale=0.17]{figures/uTsgan_mid_1000.jpg}\tabularnewline
\hline 
87.78 & 73.37 & 72.99 & 65.78\tabularnewline
\hline 
\end{tabular}
\par\end{centering}
\caption{MiddlePhalanxOutlineCorrect class 0, medium data set. Examples of spectrogram images and time series pair generated from uTSGAN for associated epochs and the FID score at each epoch. This shows uTSGAN consistently gets better at generation over training. It also depicts the dependency of the two networks, showing as the quality of the spectrogram gets better so does the time series generation.}
\label{middle_utsgan_epoch_results}
\end{figure*}

\begin{figure*}[h!]
\begin{centering}
\begin{tabular}{|c|c|c|c|}
\hline 
Epoch 250 & Epoch 500 & Epoch 750 & Epoch 1000\tabularnewline
\hline
\includegraphics[scale=0.09]{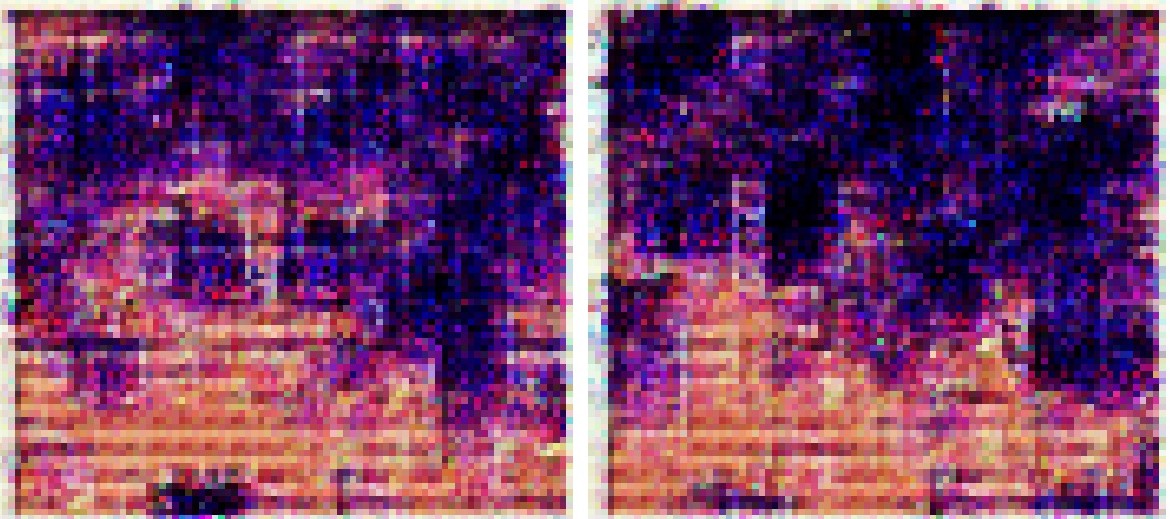} &
\includegraphics[scale=0.09]{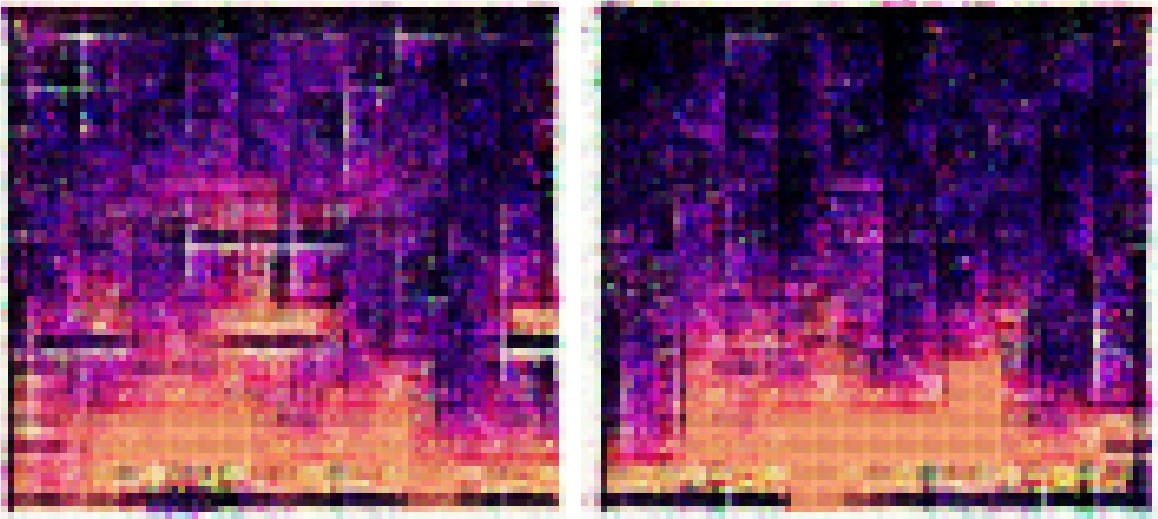} & 
\includegraphics[scale=0.09]{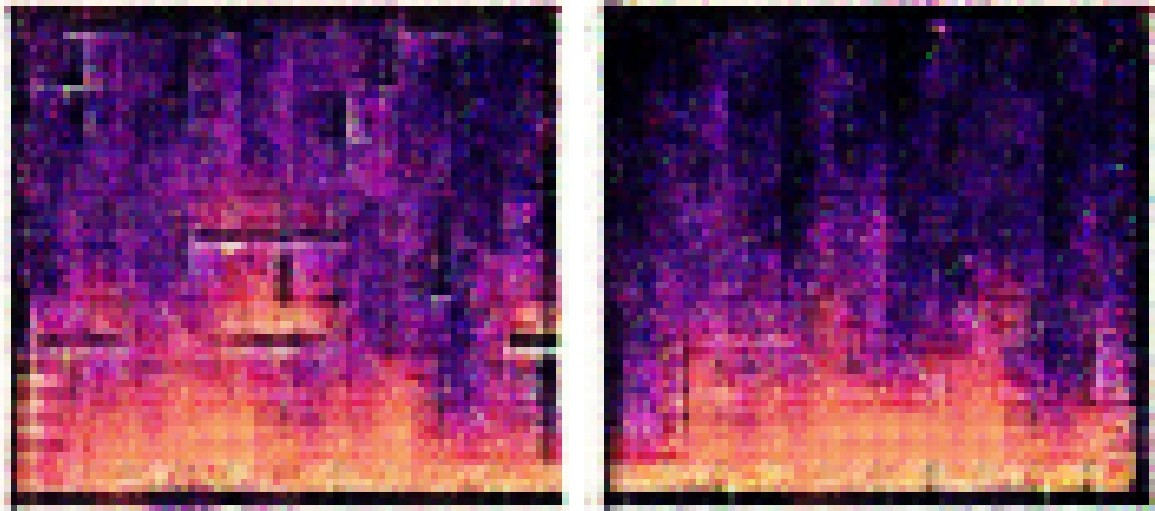} &
\includegraphics[scale=0.09]{figures/uTsgan_ham_2D1000.jpg}\tabularnewline
 
\includegraphics[scale=0.17]{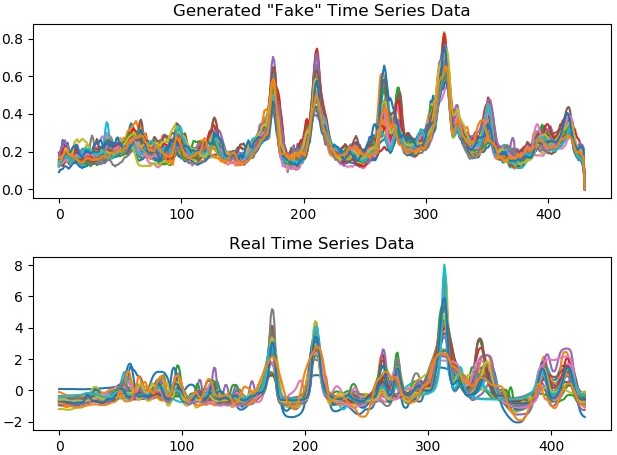} &
\includegraphics[scale=0.17]{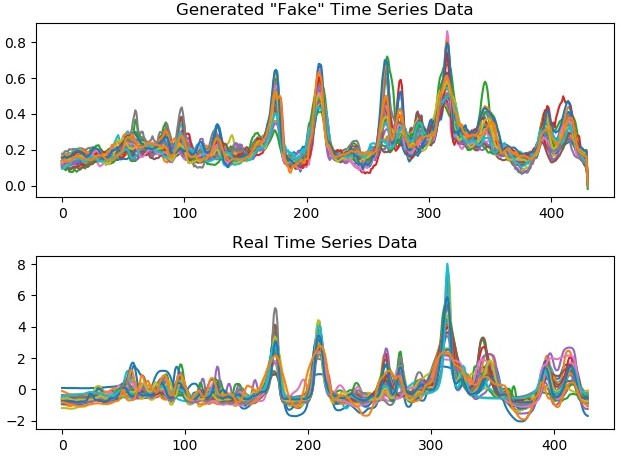} & 
\includegraphics[scale=0.17]{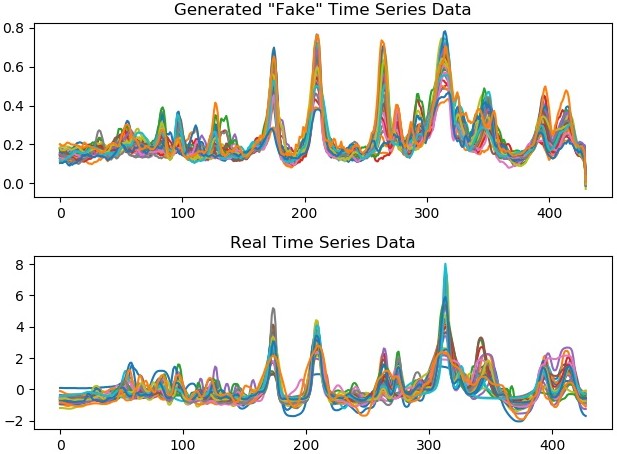} &
\includegraphics[scale=0.17]{figures/uTsgan_ham_1000.jpg}\tabularnewline
\hline 
92.52 & 88.67 & 61.16 & 26.30\tabularnewline
\hline 
\end{tabular}
\par\end{centering}
\caption{Ham class 1, small data set. Examples of spectrogram images and time series pair generated from uTSGAN for associated epochs and the FID score at each epoch. This shows uTSGAN consistently gets better at generation over training. It also depicts the dependency of the two networks, showing as the quality of the spectrogram gets better so does the time series generation.}
\label{ham_utsgan_epoch_results}
\end{figure*}

\begin{figure*}[h!]
\begin{centering}
\begin{tabular}{|c|c|c|c|}

\hline 
Epoch 250 & Epoch 500 & Epoch 750 & Epoch 1000\tabularnewline
\hline
\includegraphics[scale=0.09]{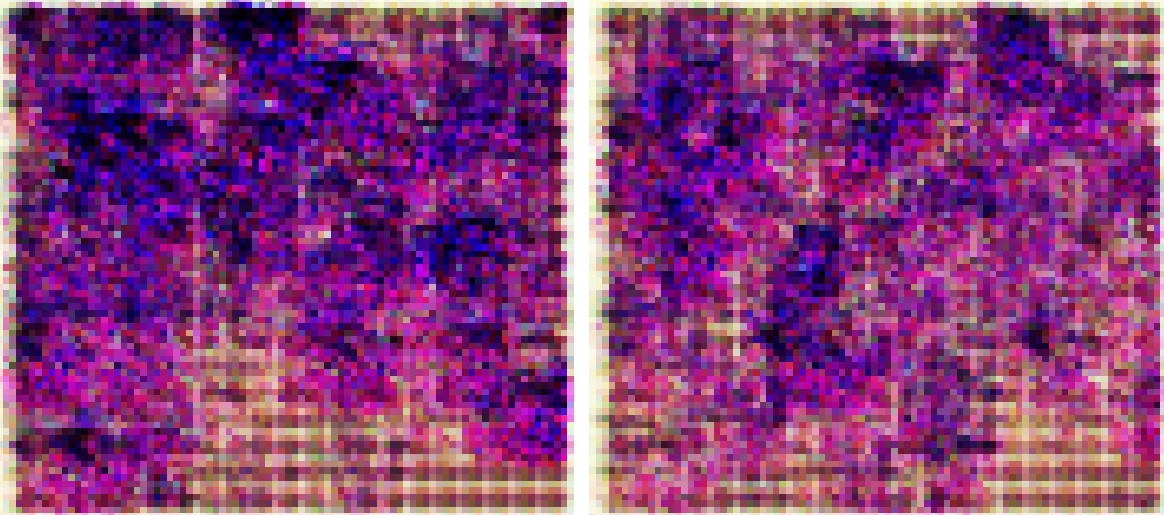} & 
\includegraphics[scale=0.09]{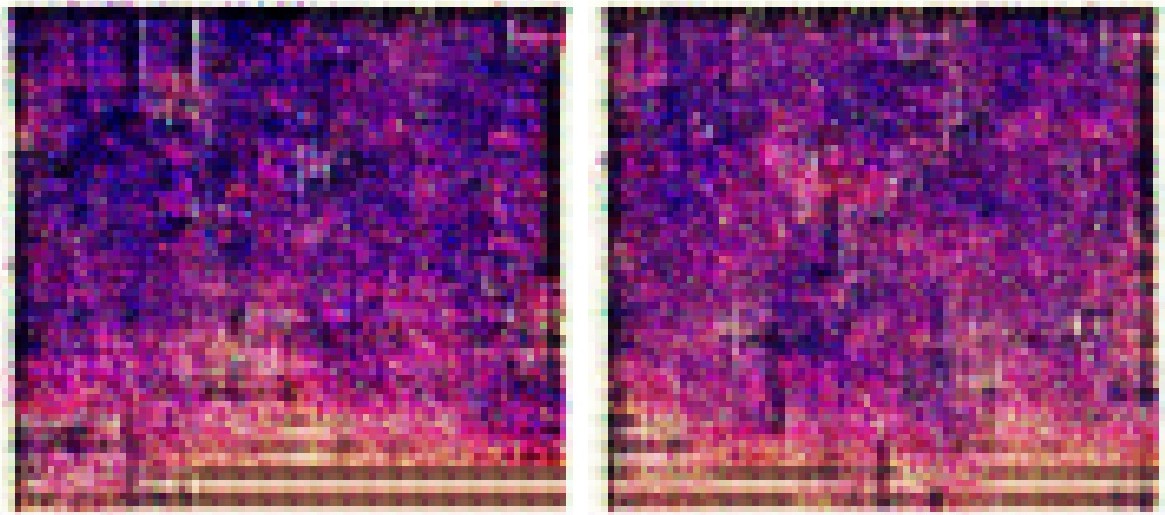} &
\includegraphics[scale=0.09]{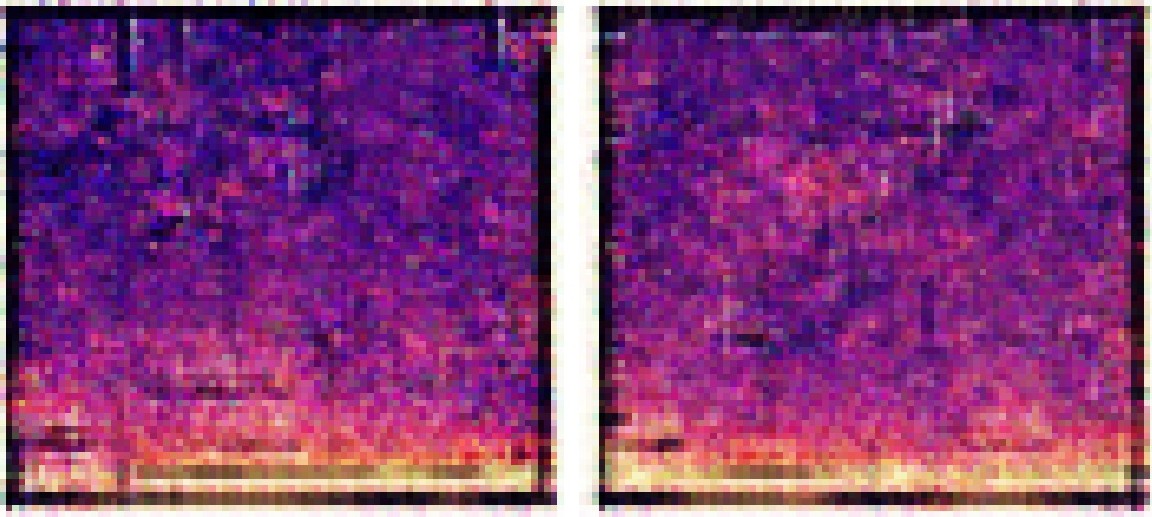} & 
\includegraphics[scale=0.09]{figures/uTsgan_beetle_2D1000.jpg}\tabularnewline
 
\includegraphics[scale=0.17]{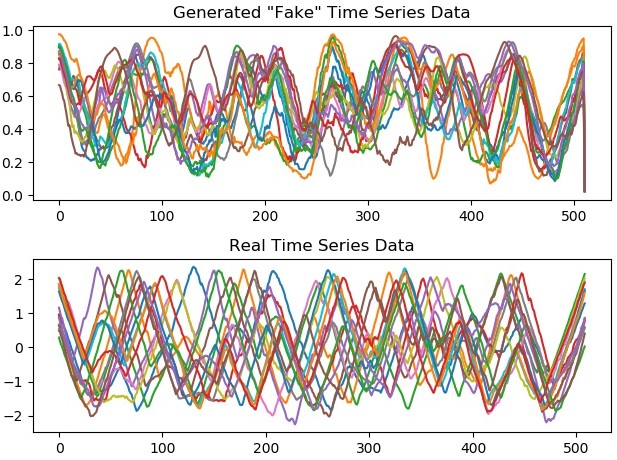} & 
\includegraphics[scale=0.17]{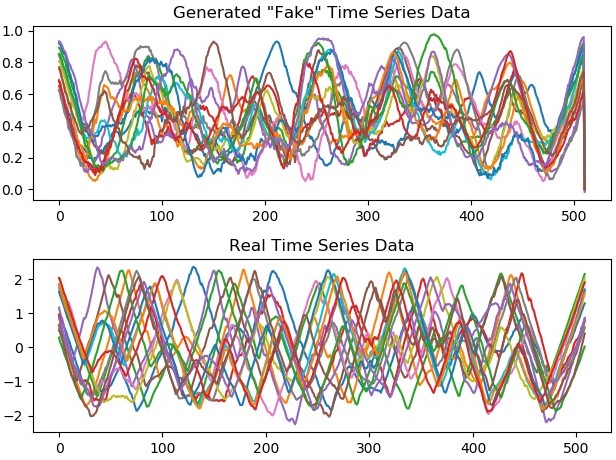} &
\includegraphics[scale=0.17]{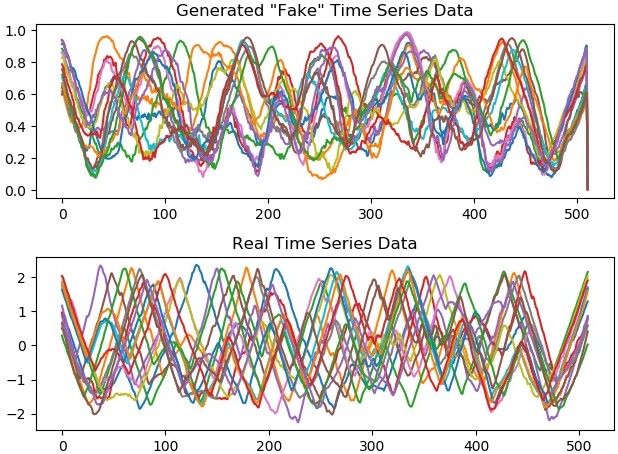} & 
\includegraphics[scale=0.17]{figures/uTsgan_beetle_1000.jpg}\tabularnewline
\hline 
48.80 & 32.33 & 14.59 & 12.61\tabularnewline
\hline 
\end{tabular}
\par\end{centering}
\caption{BeetleFly class 1, small data set. Examples of spectrogram images and time series pair generated from uTSGAN for associated epochs and the FID score at each epoch. This shows uTSGAN consistently gets better at generation over training. It also depicts the dependency of the two networks, showing as the quality of the spectrogram gets better so does the time series generation.}
\label{beetle_utsgan_epoch_results}
\end{figure*}

\begin{figure*}[h!]
\begin{centering}
\begin{tabular}{|c|c|c|c|}
\hline 
Epoch 250 & Epoch 500 & Epoch 750 & Epoch 1000\tabularnewline
\hline
\includegraphics[scale=0.09]{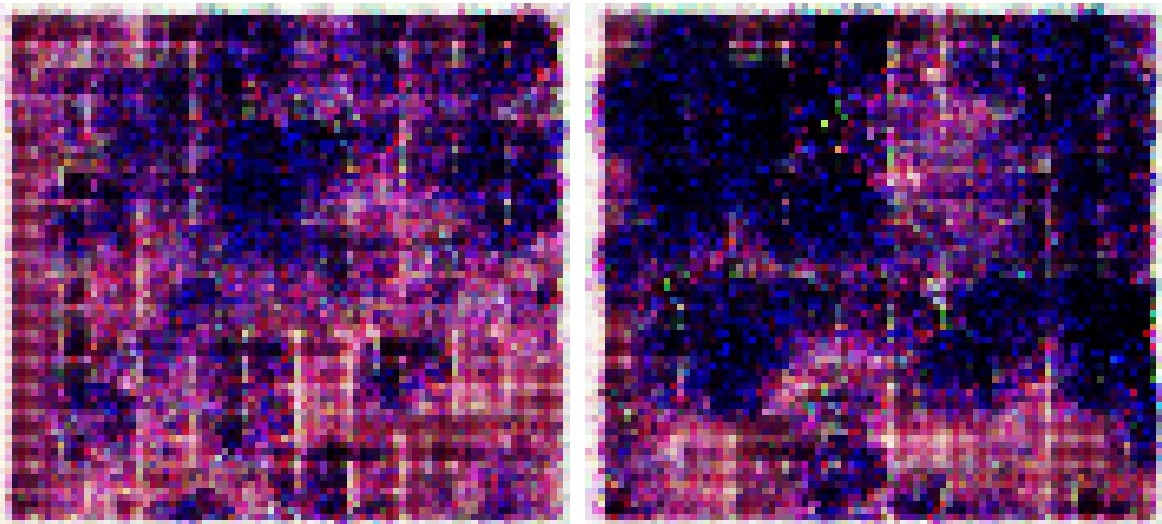} & 
\includegraphics[scale=0.09]{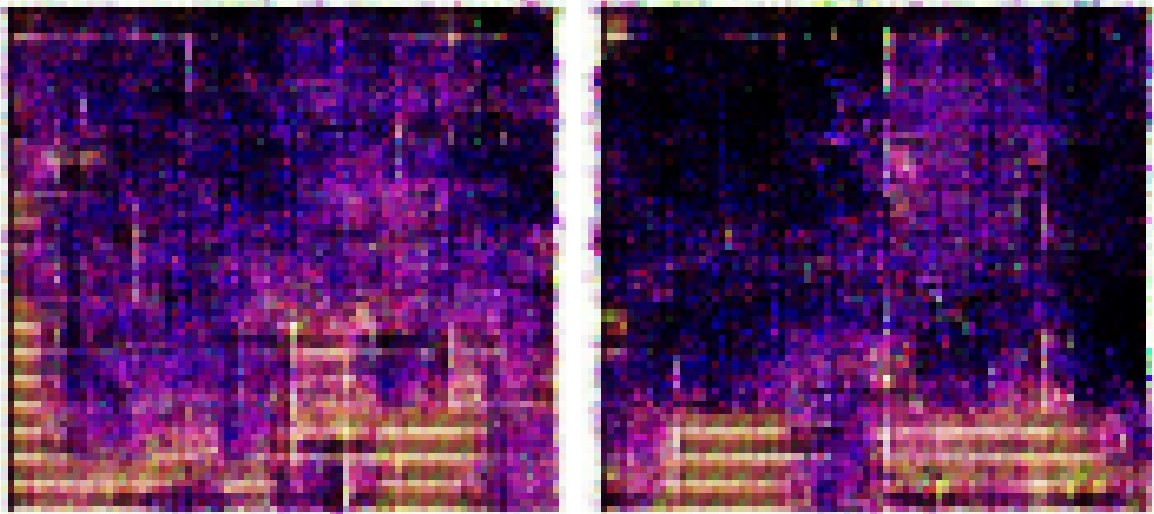} &
\includegraphics[scale=0.09]{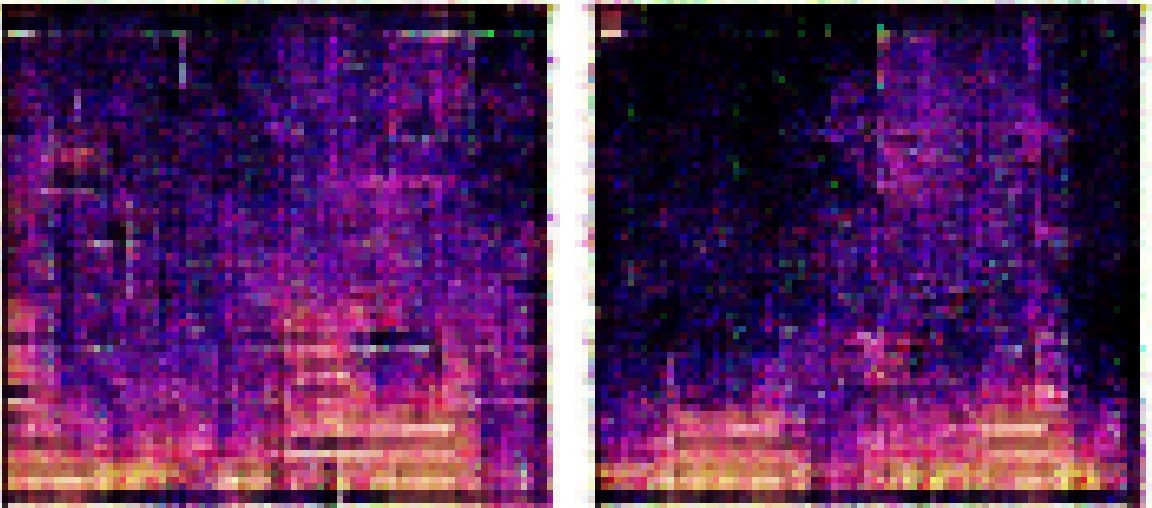} & 
\includegraphics[scale=0.09]{figures/uTsgan_beef_2D1000.jpg}\tabularnewline
 
\includegraphics[scale=0.17]{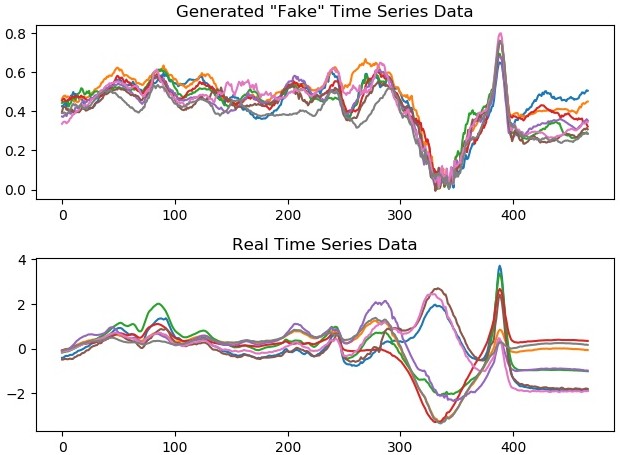} & 
\includegraphics[scale=0.17]{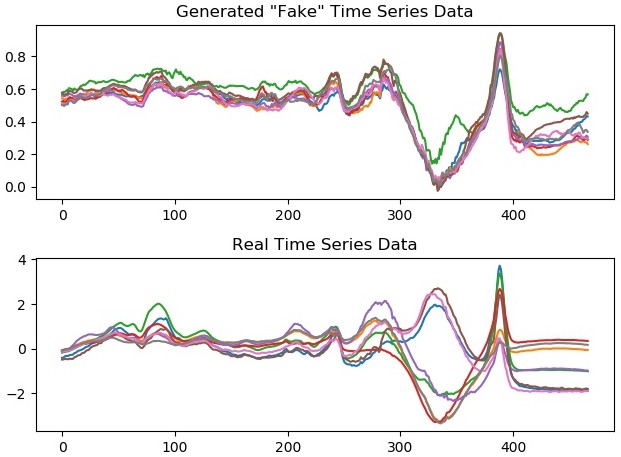} &
\includegraphics[scale=0.17]{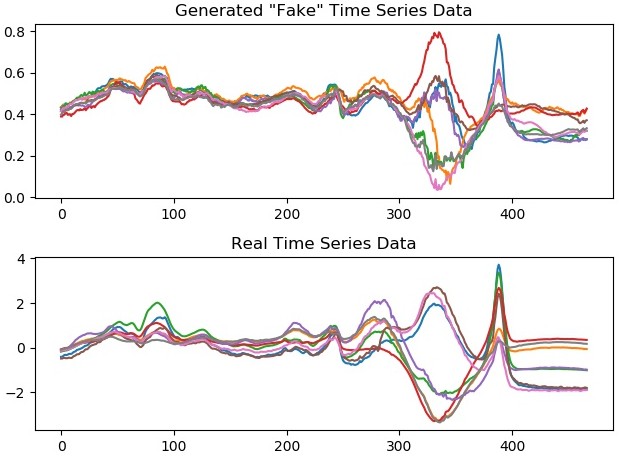} & 
\includegraphics[scale=0.17]{figures/uTsgan_beef_1050.jpg}\tabularnewline
\hline 
239.83 & 127.45 & 96.16 & 47.26\tabularnewline
\hline 
\end{tabular}
\par\end{centering}
\caption{Beef class 1,small data set.  Examples of spectrogram images and time series pair generated from uTSGAN for associated epochs and the FID score at each epoch. This shows uTSGAN consistently gets better at generation over training. It also depicts the dependency of the two networks, showing as the quality of the spectrogram gets better so does the time series generation.}
\label{beef_utsgan_epoch_results}
\end{figure*}

What we now show with our uTSGAN is the ability to have dependency across both $WGAN_{x}$ and $WGAN_{y}$ in their generation of both 2D and 1D data. Some examples can be seen more clearly in Fig \ref{yoga_utsgan_epoch_results} - Fig \ref{beef_utsgan_epoch_results} showing the generation of spectrogram images and its corresponding time series data. In this, we want to also point out the FID score is shown at the bottom for each epoch of training. More statistical
measures will be addressed regarding FID score shortly, but in Fig \ref{yoga_utsgan_epoch_results} - Fig \ref{beef_utsgan_epoch_results} it can be seen that in every data set shown uTSGAN outperforms TSGAN in FID score before the 1000th epoch of training. 

Also shown in Fig \ref{yoga_utsgan_epoch_results} - Fig \ref{beef_utsgan_epoch_results} that is interesting is the quality
of the spectrogram image being produced is extremely poor and noisy in the lower epochs of training (and not as clean even when training is complete) compared to the TSGAN's generated spectrograms in fewer epochs. Though we do realize this isn't the ideal scenario, what is shown is even when the generation of the spectrogram is poor it is still producing decent time series data. In some cases it's seen that uTSGAN generates better quality time series data in as little as 250 epochs even with the spectrogram looking indistinguishable.

What was accomplished by unifying the loss functions in TSGAN is the two sub-networks, $WGAN_{x}$ and $WGAN_{y}$ now work together to generate better time dependent data. For an abstract analogy, please follow. In the original TSGAN, the generator $G$ of $WGAN_{x}$ is like a cousin to the generator $F$ in $WGAN_{y}$ while the discriminators in both, $D_{x}$and $D_{y}$ are the bullies on the playground. Alone, $G$ can easily fend off their bully while $F$ has a hard time with theirs because $D_{y}$ is a much larger more dangerous bully. Once $G$ is done with $D_{x}$ they help out $F$ but from behind the fence because it's not really their fight (i.e. shouting words of encouragement, or telling them how they dealt with their bully, etc.). 

Now, with uTSGAN, $G$ takes the role of an older sibling instead of cousin, one that is more interested in helping $F$ fend off $D_{y}$ instead of just handling $D_{X}$ and there is no fence. Together, $G$ and $F$ help each other out, since $D_{y}$ is still the greater danger more emphasis is taken into dealing with that bully by both $G$ and $F$. While $D_{x}$ is dealt with in turn once $D_{y}$ is at bay momentarily. This metaphor of big sibling helping little sibling and a two-on-two scenario versus a one-on-one is seen in Fig \ref{yoga_utsgan_epoch_results} - Fig \ref{beef_utsgan_epoch_results} as well. The generator $G$ allows for its spectrograms to fault during training in order to help out the generation of $F$'s time series data. Though $D_{X}$ can distinguish $G$'s spectrograms from real ones easily alone, when $D_{x}$ is dealing with the loss of $D_{y}$ as well it allows $F$'s generation to happen better and
quicker. 

\subsection{Quantitative Analysis}
To reiterate, our experiments are over 70 different univariate data sets from the UCR's time series archive. We now want to look at how well our uTSGAN performs on these data sets across multiple ran experiments and looking at different epoch outputs during training. During training, uTSGAN's model and weights are saved at 250, 500, 750, and 1000 epochs
(which is the end of our training). Once training is complete, we train an FCN on the original data set for 1000 epochs or until convergence through validation accuracy and we use this trained FCN network now to calculate our FID scores. Each uTSGAN model (250, 500, 750, 1000) is ran to generate synthetic data and then that data is used in comparison to the real data to determine the realism of the generated samples.  FID scores were calculated in this fashion 25 times feeding different random vectors to $WGAN_{x}$ in order to generate different time series data in $WGAN_{y}$. Again, the FID scores in this paper are the average FID scores throughout these 25 different runs of the generation. 

With the 70 data sets, there are 28 data sets that are seen as small in training samples, 26 that are medium, and 16 that are large. For the delineation of small, medium, and large, we refer readers to subsection \ref{subsec:Data} for reference. At a high level, uTSGAN outperformed TSGAN in 55 of the 70 data sets by the 1000th epoch of training. In fact, uTSGAN achieved a lower FID score after 750 epochs of training in 43 of the 70 data sets. This alone shows empirically that uTSGAN trains better than that of TSGAN and converges to better generated time series quicker. 

Breaking down the data sets by training size, the small training sets, uTSGAN outperforms TSGAN in FID score for only four of the data sets by the 250th epoch, 11 data sets by the 500th epoch, 16 data sets at 750th epochs, and 22 data sets by the 1000th epoch. This shows over the course of training uTSGAN is continuously getting better at generation. Another interesting statistic is when uTSGAN outperforms the 22 data sets in the small category, it does so by an average of
2.94 times better than TSGAN while in those six data sets in which uTSGAN never performed better than it was on average 0.36 times worse.  This could be contributed to needing more time for the training, i.e. more epochs; but is more likely the case that the generated spectrograms are of poor quality compared to the time series contributed and that penalty in the loss function overwhelmed the time series. 

For the medium training data sets, uTSGAN achieves better FID scores than ten data sets by the 250th epoch, twelve by the 500th epoch, 19 by the 750th epoch, and 22 by the end of training. Of those four data sets not achieving a better FID, uTSGAN was only 0.63 times worse than TSGAN while being 6.93 times better in those data sets it beat.  This again shows progression of learning over time showing an increase in performance while training continues.

A similar story is seen when looking at those data sets classified as large training. uTSGAN outperforms four of the data sets by the 250th epoch, five by the 500th epoch, eight by the 750th epoch, and eleven by the end of training. Of those data sets which uTSGAN did not perform better, our FID score was only 0.39 times worse than that of TSGAN; while the winning data sets were 8.23 times better in FID scores. 

Overall uTSGAN achieves FID scores 6.04 times better than the original TSGAN, showing the unified loss function and dependency in training is a value added to the time series generation. It also is important to note that during training. all data sets saw an increase in FID performance as training went on, showing consistent learning and generation happening in this unified architecture of sibling GANs. 

%% file: Conclusion.tex
\section{Conclusion}
\label{conc}

The possibilities of solving rare, complex problems are endless if
there was an abundance of data for each issue. If a researcher was
able to supply an algorithm with enough data, the chance of solving
large problems increase; e.g. forecasting weather patterns, predicting
the stock exchange, identifying what a rare heart disease looks like
in an ECG, or catching a brain tumor in a EEG in the early stages.
As can be imagined, some of these problems can't supply a researcher
with enough data and doing other means to collect the data is difficult.
That is why data generation is such an important task in the machine
learning community and supports large interest with researchers world
wide. These events described above are time series dependent events
and it happens that time series data is one of the more complicated
types of data to generate. 

It wasn't until recently that the first few shot approach, TSGAN,
was published by Smith et al. that showed promise for data generation.
In this paper though we express our concerns with TSGAN and address
them accordingly to create a novel mathematical approach to similar
few shot learning algorithm, uTSGAN. uTSGAN uses the same time frequency
information which showed to help TSGAN in time dependent signal generation,
but yet increases its ability by unifying both sub-networks inside
the architecture. With this unification, we have shown uTSGAN can
generate realistic time series on 70 public data sets standard to
the time series community, spanning different tasks and collection
modalities. We show our approach can generate data faster than TSGAN,
achieving a higher FID score in more than 60\% of the data sets compared
in 75\% of the training time needed. We also have shown that uTSGAN
generates better quality time series over the same amount of training
time, resulting in an FID score six times better than that of TSGAN
in over 75\% of the data compared. 

Though we mention that the data used for uTSGAN generation have such
small training size that all the data sets used could be considered
\emph{few shot}, we delineate the small training data sets of having
less than 500 signals in their training. From these 28 small size
data sets and 26 medium size one (all 54 data sets have less than
1000 training signals), uTSGAN outperforms TSGAN on over 80\% of the
data sets tested by a combined factor almost five times better. This
result shows the \emph{few shot} ability of uTSGAN is preserved with
this unification. 

\subsection{Future Work}

Although we have shown some great progress in uTSGAN for time series
generation, we definitely do not think we solved the problem of a
universal generation method. There are several things we would like
to advance in the generation ability and hope the community finds
this work essential enough to advance as well. First, there needs
to be a better way of quantitatively verifying the realism of the
time series data being generated. The original paper for TSGAN classification
metrics that show promise as well as 1D FID score, but there needs
to be a universal way to generate any time series data and then compare
to the original time series which meets standardization in the community. 

As far as work on uTSGAN, we would like to see if there is any benefit
for many of the training ``tricks'' used in image generation. Things
like pixel normalization (of course mapped to 1D), progressively growing
the signal, more conditioning of the input with actual spectral information
and not just RGB values, injecting information in the layers while
learning is going on \cite{Karras2017ProgressiveVariation,Karras2019ANetworks,Donahue2018AdversarialSynthesis}. Though these things are not of interest for this paper,
it would be great to see application-based papers coming forward and
finding tricks like these or different ones that push the
data generation one step closer to realism. Also, we see the ability to add some auxiliary classifier to the end of the discriminator as well as conditioning the input of the generator into multi-class label approach to extend the generation task further. There needs to be some
addressing of the noise added in during the training and how to smooth
that out through generation. It could be that a simple filter at the
end of the network would work or adding some technique like WaveGAN's
phase shuffling for audio samples \cite{Donahue2018AdversarialSynthesis}.

We would also like to explore the unified loss function itself and
see if penalizing the spectrogram image generation as well as the
reconstruction of the time series data from the image can help in
our task. Our interest in this goes back to Section \ref{expres} Fig \ref{yoga_utsgan_epoch_results} - \ref{beef_utsgan_epoch_results}, showing that as the spectrogram image begins to get more
prominent and resemble the original spectrogram, the time series generation
also increases. We saw uTSGAN show a better FID score as training
continued and we expect this spectral conditioning and dependency
now holds a key to unlocking better time series generation. 

Lastly, we look to other approaches of not just our WGAN approach.
With this unification, it is possible some of these GAN loss function
with actual mathematical convergence could make generation easier,
better, and faster than that of WGAN. We open up the research community
to look into other architectural losses and networks in uTSGAN to
extend the work further to come one algorithm closer to a universal
time series generation method.